\documentclass[twoside,12pt]{article}

\usepackage[margin=1in]{geometry} 

\usepackage[utf8]{inputenc}
\usepackage{amsfonts}
\usepackage{mathrsfs}
\usepackage{dsfont}
\usepackage{amsmath,amssymb}
\usepackage{latexsym}
\usepackage{mathtools} 

\usepackage{graphicx}
\usepackage{float}
\usepackage{enumitem}
\usepackage{longtable}
\usepackage{booktabs}
\usepackage{epstopdf}
\usepackage{tabularx}
\usepackage{multirow}
\usepackage{siunitx}
\usepackage{makecell}      
\usepackage{subcaption}    

\usepackage{tikz}
\usetikzlibrary{shapes.geometric, arrows, positioning}
\tikzstyle{startstop} = [rectangle, rounded corners, minimum width=3cm, minimum height=1cm, text centered, draw=black, fill=red!30]
\tikzstyle{process} = [rectangle, minimum width=3cm, minimum height=1cm, text centered, draw=black, fill=blue!30]
\tikzstyle{io} = [trapezium, trapezium left angle=70, trapezium right angle=110, minimum width=3cm, minimum height=1cm, text centered, draw=black, fill=green!30]
\tikzstyle{arrow} = [thick,->,>=stealth]

\usepackage{natbib}
\usepackage{algorithm}
\usepackage{algpseudocode} 

\usepackage{listings}
\usepackage{xcolor}
\definecolor{codegreen}{rgb}{0,0.6,0}
\definecolor{codegray}{rgb}{0.5,0.5,0.5}
\definecolor{codepurple}{rgb}{0.58,0,0.82}
\definecolor{backcolour}{rgb}{0.95,0.95,0.92}
\lstdefinestyle{mystyle}{
    backgroundcolor=\color{backcolour},
    commentstyle=\color{codegreen},
    keywordstyle=\color{magenta},
    numberstyle=\tiny\color{codegray},
    stringstyle=\color{codepurple},
    basicstyle=\footnotesize\ttfamily,
    breakatwhitespace=false,
    breaklines=true,
    captionpos=b,
    keepspaces=true,
    numbers=left,
    numbersep=5pt,
    showspaces=false,
    showstringspaces=false,
    showtabs=false,
    tabsize=2
}
\usepackage[
    bookmarks=true,
    bookmarksnumbered=true,
    colorlinks=true,
    pdfstartview=FitV,
    linkcolor=blue,
    citecolor=blue,
    urlcolor=blue
]{hyperref}
\usepackage{fancyhdr}

\definecolor{Brown}{rgb}{0.64,0.16,0.16}
\definecolor{OliveGreen}{rgb}{0.1,0.4,0.1}

\newtheorem{theorem}{Theorem}

\newenvironment{proof}[1][Proof]{\textbf{#1.} }{\ \rule{0.5em}{0.5em}}

\begin{document}

\title{\Large \textbf{IGNIS: A Robust Neural Network Framework for Constrained Parameter Estimation in Archimedean Copulas}}

\author{Agnideep Aich\thanks{Corresponding author: \texttt{agnideep.aich1@louisiana.edu}} \\ 
        Department of Mathematics \\
        University of Louisiana at Lafayette \\
        Lafayette, Louisiana, USA}
\date{} 
\maketitle

\begin{abstract}
Classical estimators, the cornerstones of statistical inference, face insurmountable challenges when applied to important emerging classes of Archimedean copulas. These models exhibit pathological properties, including numerically unstable densities, a restrictive lower bound on Kendall's tau, and vanishingly small likelihood gradients, making MLE brittle and limiting MoM’s applicability to datasets with sufficiently strong dependence (i.e., only when the empirical Kendall’s $\tau$ exceeds the family’s lower bound $\approx 0.545$). We introduce \textbf{IGNIS}, a unified neural estimation framework that sidesteps these barriers by learning a direct, robust mapping from data-driven dependency measures to the underlying copula parameter $\theta$. IGNIS utilizes a multi-input architecture and a theory-guided output layer ($\mathrm{softplus}(z) + 1$) to automatically enforce the domain constraint $\hat{\theta} \ge 1$. Trained and validated on four families (Gumbel, Joe, and the numerically challenging A1/A2), IGNIS delivers accurate and stable estimates for real-world financial and health datasets, demonstrating its necessity for reliable inference in modern, complex dependence models where traditional methods fail. To our knowledge, IGNIS is the first \emph{standalone, general-purpose} neural estimator for Archimedean copulas (not a generative model or likelihood optimizer), delivering direct, constraint-aware $\hat\theta$ and readily extensible to additional families via retraining or minor output-layer adaptations.

\end{abstract}

\noindent {\bf Keywords:}
A1 and A2 copulas, Archimedean copulas, Parameter estimation, Neural networks, Computational statistics
\medskip

\section{Introduction}
\label{sec:introduction}

Maximum Likelihood Estimation (MLE), a pillar of statistical inference, is the gold standard for parameter estimation due to its desirable asymptotic properties. Its efficacy, however, is predicated on well-behaved likelihood functions. In the domain of dependence modeling using copulas \citep{nelsen2006introduction}, this assumption can dramatically fail. For a growing class of flexible and important models, such as the novel A1 and A2 Archimedean copulas \citep{aich2025two}, the likelihood function exhibits pathological properties that render classical estimation methods inconsistent, unstable, or computationally infeasible. This issue is not isolated; numerical challenges in copula estimation are a known and significant concern in high-stakes applications like quantitative risk management \citep{hofert2013archimedean}.

In this paper we focus on the \emph{practical} regime $\theta\in[1,20]$. Our aim is not to make an asymptotic case against classical methods, but to document and address the \emph{finite-sample brittleness} we observe in this range and to provide a robust alternative. Three facts motivate our approach—two are empirical, and one is a diagnostic explanation:
\begin{enumerate}
    \item \textbf{Finite-range brittleness of likelihood-based estimators.} In a controlled MLE/MPL stress test for A1/A2 within $\theta\in[1,20]$, we observe optimizer failures on stabilized objectives, mask-sensitive estimates, and non-smooth objectives (“kinks”) at higher $\theta$—all despite careful numerical safeguards.
    \item \textbf{Applicability gap for MoM.} For the Method of Moments (MoM) to be viable, a model’s theoretical dependence range must cover the data’s empirical dependence. A1 and A2 are severely constrained here: their Kendall’s $\tau$ begins at  $\approx 0.54518$, so MoM is \emph{available} only when the empirical $\tau \geq 0.545$.
    \item \textbf{Diagnostic asymptotics.} As a \emph{diagnostic} (not the main result), we show that even moderate $\theta$ values can produce flat likelihood surfaces: scores and Hessians decay with $\theta$, clarifying why optimizers stall or become sensitive in practice.
\end{enumerate}

Recent deep learning approaches have shown immense promise in statistics, but have not addressed this specific estimation problem. The state of the art has largely focused on generative tasks, such as learning new copula generators from scratch \citep{ling2020deep, ng2021generative} or modeling highly complex, high-dimensional dependence structures \citep{ng2022archimax}. However, the fundamental problem of robust \emph{parameter estimation} for known, specified families that exhibit the aforementioned pathologies remains a critical open gap. To fill this gap, we introduce \textbf{IGNIS}, a unified neural estimation framework that sidesteps the pitfalls of classical methods entirely.

\[
\underbrace{\text{Classical copula estimation}}_{\substack{\text{MoM, MLE, MPL}\\\text{Pathological Failures}}}
\;\longrightarrow\;
\underbrace{\text{Existing Neural Copulas}}_{\substack{\text{Generative Focus}\\\text{No Parameter Estimation Tool}}}
\]
\[
\quad\longrightarrow\quad
\underbrace{\textbf{This Work}}_{\substack{\text{IGNIS: Robust, Unified}\\\text{Parameter Estimator}}}
\]

\noindent\textbf{Positioning.} IGNIS is a \emph{standalone, discriminative} estimator: given a small set of robust dependence summaries and a family tag, it returns a constraint-respecting $\hat\theta$ without evaluating densities or optimizing likelihoods. It works for Gumbel/Joe as well as A1/A2 and is readily extensible to other families.

IGNIS learns a direct mapping from a vector of robust, data-driven summary statistics to the underlying copula parameter $\theta$. Our main contributions are:
\begin{enumerate}
    \item \textbf{IGNIS: a general-purpose, constraint-aware estimator.} We propose a unified neural architecture that maps features $\rightarrow$ parameter directly and enforces $\hat{\theta} \ge 1$ via a softplus$+1$ output layer, avoiding brittle likelihoods while supporting multiple families.
    \item \textbf{Practical validation in the finite range.} We provide a systematic simulation study across four families (Gumbel, Joe, A1, A2) and evaluate out-of-sample performance against MoM using paired $t$ and Wilcoxon signed-rank tests, equivalence via TOST at $\varepsilon=10^{-3}$ nats/obs, and bootstrap standard errors.
    \item \textbf{Why classical methods struggle here.} We document finite-range MLE/MPL brittleness via a stress test and supply asymptotic diagnostics (score/Hessian decay) that explain the observed optimizer behavior.
\end{enumerate}

\emph{Scope.} We do not advocate fitting A1/A2 when $\tau_n<0.545$; rather, when such families are selected (e.g., by prior analysis suggesting strong joint tails) and classical estimators are brittle or unavailable, IGNIS provides a stable parameter estimate while also functioning for standard families like Gumbel and Joe.

The remainder of this paper is organized as follows. Section~\ref{sec:relatedwork} reviews related work. Section~\ref{sec:notation} presents the notations used in the paper. Section~\ref{sec:background} presents necessary preliminaries. Section~\ref{sec:motivation} provides motivation for our work. Section~\ref{sec:methods} details the IGNIS architecture and training protocol. Section~\ref{sec:simulation} presents the simulation results for IGNIS, and Section~\ref{sec:applications} demonstrates real-data applications. Finally, Section~\ref{sec:conclusion} concludes and outlines future research directions.

\section{Related Work}
\label{sec:relatedwork}

Our work builds upon two distinct streams of literature: classical parameter estimation for copulas and the emerging field of deep learning for statistical modeling.

\subsection{Classical Estimation and its Limitations}

Parameter estimation for Archimedean copulas has traditionally been approached via two main routes. The Method of Moments (MoM), particularly using Kendall's $\tau$ or Spearman's $\rho$, is valued for its computational simplicity and circumvention of the likelihood function \citep{genest1993statistical}. However, both A1 and A2 have a \textbf{high lower bound} for Kendall's $\tau$ ($8\ln 2 - 5 \approx 0.54518$), which makes MoM inapplicable to many real datasets with weaker dependence.

The second route is Maximum Likelihood Estimation (MLE) or its semi-parametric variant, Maximum Pseudo-Likelihood (MPL) \citep{genest1995semiparametric}. While asymptotically efficient, MLE requires computing the copula density, which can be analytically complex and numerically unstable. Efforts by researchers \citep{hofert2012likelihood} derived explicit generator derivatives to make MLE more feasible for standard families. Yet, subsequent large-scale studies confirmed that even with these advances, classical estimators face significant numerical challenges and potential unreliability, especially in high dimensions or for complex models \citep{hofert2013archimedean}. The A1 and A2 families are prime examples where these numerical pathologies become insurmountable barriers, necessitating a new approach.

\subsection{Deep Learning Approaches to Copula Modeling}

The recent intersection of deep learning and copula modeling has been dominated by powerful generative approaches that learn or approximate the generator function itself, rather than estimating parameters of a pre-defined family. For instance, ACNet \citep{ling2020deep} introduced a neural architecture to learn completely monotone generator functions, enabling the approximation of existing copulas and the creation of new ones. Similarly, \citep{ng2021generative} proposed a generative technique using latent variables and Laplace transforms to represent Archimedean generators, scaling to high dimensions. Other work has focused on non-parametric inference for more flexible classes like Archimax copulas, which are designed to model both bulk and tail dependencies \citep{ng2022archimax}.

While these methods represent the state-of-the-art in constructing flexible, high-dimensional dependence models, they do not address the targeted problem of estimating the parameter $\theta$ for a specified family, especially when that family exhibits the estimation pathologies we have identified. Broader work on Physics-Informed Neural Networks (PINNs) has shown the power of deep learning for solving problems with known physical constraints \citep{raissi2019physics, sirignano2018dgm}, but a specialized framework for constrained parameter estimation in statistically challenging copula models has been a missing piece.  In contrast, IGNIS targets the missing piece: \emph{parameter estimation} for specified families. It neither learns a generator nor maximizes a brittle likelihood; it provides a direct, reusable estimator with calibrated uncertainty via bootstrap. IGNIS is designed specifically to fill this gap, providing a discriminative estimator that is robust, constraint-aware, and applicable across multiple families where classical methods fail. Even for families without a $\tau$ floor (e.g., Gumbel, Joe), IGNIS removes likelihood-evaluation fragility and the need for closed-form inversions, delivering fast, stable point estimates with straightforward bootstrap uncertainty. Thus, the framework is useful beyond the A1/A2 motivation and complements classical methods even when they are available.

\section{Notation}
\label{sec:notation}

Throughout our analysis, we employ a consistent set of symbols. The core parameter of an Archimedean copula is denoted by $\theta \in [1, \infty)$, with its estimate from our framework being $\hat{\theta}$. The copula function itself is $C(u,v)$, constructed via a generator function, $\phi(t)$, and its inverse, $\phi^{-1}(s)$. To ensure clarity, we distinguish this from its corresponding probability density function, $c(u,v)$. In theoretical contexts (Appendix~\ref{app:Appendix C}), the standalone uppercase letter $C$ denotes the copula family, while subscripted variants (e.g., $C_k$) represent constants within proofs. For the Method of Moments, we use the theoretical Kendall's tau, denoted by $\tau$.

Our neural network framework, IGNIS, is trained on a dataset of $N$ examples. Each example is an input vector $\mathbf{x} \in \mathbb{R}^9$. This vector is a concatenation of two components: a 5-dimensional vector of continuous summary features, $\mathbf{f} \in \mathbb{R}^5$, and a 4-dimensional one-hot vector indicating the copula family, $\mathbf{c} \in \{0,1\}^4$. The feature vector $\mathbf{f}$ is comprised of five empirical dependency measures calculated from a data sample of size $n$: Kendall's tau ($\tau_n$), Spearman's rho ($\rho_n$), the Pearson correlation coefficient ($r_n$), and coefficients of upper ($\lambda_{upper,n}$) and lower ($\lambda_{lower,n}$) tail dependence.

The neural network has $D$ layers and is trained to minimize a mean squared error loss function $\mathcal{L}(\theta)$ by adjusting its weights and biases using the Adam optimizer with a learning rate $\eta$. For the theoretical consistency proof presented in Appendix~\ref{app:Appendix C}, the feature vector is denoted by $\mathbf{T}_n$, and the set of all possible feature vectors is the feature space $\mathcal{T}$.
\section{Preliminaries}
\label{sec:background}

\subsection{Copulas and Dependency Modeling}
Copulas are statistical tools that model dependency structures between random variables, independent of their marginal distributions. Introduced by  \citet{sklar1959fonctions}, they provide a unified approach to capturing joint dependencies. Archimedean copulas, known for their simplicity and flexibility, are defined using a generator function, making them particularly effective for modeling bivariate and multivariate dependencies.

\subsection{The A1 and A2 Copulas}

Like all Archimedean copulas, the novel A1 and A2 copulas \citep{aich2025two} are defined through generator functions $\phi(t)$ that are continuous, strictly decreasing, and convex on $[0,1]$, with $\phi(1)=0$. The A1 and A2 copulas extend the Archimedean copula framework to capture both upper and lower tail dependencies more effectively. In general, an Archimedean copula is given by:
\begin{equation}
C(u,v) = \phi^{-1}\bigl(\phi(u)+\phi(v)\bigr).
\end{equation}
For the A1 copula, the generator and its inverse are defined as:
\begin{align}
\phi_{A1}(t;\theta) &= \Bigl(t^{1/\theta}+t^{-1/\theta}-2\Bigr)^{\theta}, \quad \theta\ge1, \\
\phi_{A1}^{-1}(t;\theta) &= \left[\frac{t^{1/\theta}+2-\sqrt{\Bigl(t^{1/\theta}+2\Bigr)^2-4}}{2}\right]^{\theta}, \quad \theta\ge1.
\end{align}
Similarly, for the A2 copula:
\begin{align}
\phi_{A2}(t;\theta) &= \Bigl(\frac{1-t}{t}\Bigr)^{\theta}(1-t)^{\theta}, \quad \theta\ge1, \\
\phi_{A2}^{-1}(t;\theta) &= \frac{t^{1/\theta}+2-\sqrt{\Bigl(t^{1/\theta}+2\Bigr)^2-4}}{2}, \quad \theta\ge1.
\end{align}
The exact formula of the Kendall's $\tau$ for A1 and A2 copulas are given by (See Appendix~\ref{app:Appendix A} for full derivations)
\begin{align}
\tau_{A1} &= 3 + 4\theta\left[\psi(\theta) - \psi\left(\theta+\frac{1}{2}\right)\right],\\
\tau_{A2}
&= 1 - \frac{6 - 8\ln 2}{\theta}\,.
\end{align}

While Eq. 6 is complex, it can be shown that $\tau_{A1}(\theta)$ is strictly monotone increasing on its entire domain of $\theta \geq 1$; a formal proof is provided in the Appendix~\ref{app:Appendix A}.

Both copulas are parameterized by $\theta\ge1$, which governs the strength and nature of the dependency. The dual tail-dependence structure of A1 and A2 copulas is particularly valuable for modeling extreme co-movements in joint distributions. In financial risk management, they can capture simultaneous extreme losses (lower-tail) and windfall gains (upper-tail), improving estimates of portfolio tail risk. In anomaly detection, they identify coordinated extreme events (e.g., simultaneous sensor failures in industrial systems or cyber attacks across networks) by quantifying asymmetric tail dependencies. This flexibility makes them superior to single-tailed copulas e.g., Clayton (captures only lower tails) and Gumbel (captures only upper tail) in scenarios where both tail behaviors are critical.

\medskip
\noindent \emph{Note}: In \citet{aich2025two}, the copulas were denoted by A and B. We refer to them as A1 and A2 in this paper.

\subsection{Simulation from Archimedean Copulas}
\label{sec:Algo}

In this section, we present an algorithm introduced by \citet{genest1993statistical} to generate an observation \((u,v)\) from an Archimedean copula \(C\) with generator \(\phi\).

\begin{algorithm} [htpb]
\caption{Bivariate Archimedean Copula Sampling \citep{genest1993statistical}} 
\label{alg:archimedean_sampling}
\begin{algorithmic}[1]
  \Require Generator \(\phi\), its derivative \(\phi'\), inverse \(\phi^{-1}\)
  \Ensure A single draw \((u,v)\) from the copula
  \State Draw \(s,\,t \overset{\text{iid}}{\sim} \mathrm{Uniform}(0,1)\)
  \State Define 
    \[
      K(x) \;=\; x \;-\;\frac{\phi(x)}{\phi'(x)}\,, 
      \quad
      K^{-1}(y) \;=\;\sup\{\,x \mid K(x)\le y\}
    \]
  \State Compute \(w \gets K^{-1}(t)\)
  \State Compute 
    \[
      u \;\gets\;\phi^{-1}\bigl(s\,\phi(w)\bigr), 
      \quad
      v \;\gets\;\phi^{-1}\bigl((1-s)\,\phi(w)\bigr)
    \]
  \State \Return \((u,v)\)
\end{algorithmic}
\end{algorithm}

The above algorithm is a consequence of the fact that if \( U \) and \( V \) are uniform random variables with an Archimedean copula \(C\), then \( W = C(U,V) \) and \( S = \frac{\phi(U)}{\phi(U) + \phi(V)} \) are independent, \( S \) is uniform \((0,1)\), and the distribution function of \( W \) is \( K \). In our implementation, the inverse function \(K^{-1}(y)\) is computed numerically using a robust root-finding algorithm (specifically, the bisection method).

\subsection{Method of Moments Estimation}
The Method of Moments (MoM) is a classical statistical technique for parameter estimation, where theoretical moments of a distribution are equated with their empirical counterparts. In the context of copula modeling, MoM is particularly advantageous when direct likelihood-based estimation is challenging due to the complexity of deriving tractable probability density functions.

In this work, we derive exact analytical formulas for Kendall's $\tau$ for both A1 and A2 copulas (see Appendix~\ref{app:Appendix A}). These formulas establish a direct relationship between Kendall's $\tau$ and the copula parameter $\theta$, allowing for robust parameter estimation. By inverting this relationship, we develop MoM estimators for $\theta$, providing a practical approach for modeling dependencies in scenarios where traditional methods like MLE and MPL may be ineffective. However, in Section~\ref{sec:motivation}, we see that for both A1 and A2, MoM is not efficent.

\section{Motivation}
\label{sec:motivation}
\subsection{Limitations in Parameter Estimation Using Method of Moments}

The Method of Moments (MoM), which works by inverting a measure of dependence like Kendall's $\tau$, is a cornerstone of classical estimation. However, its use is predicated on a simple condition: the theoretical range of a copula family's $\tau$ must be able to represent the empirical $\tau$ calculated from a dataset. For many common families, this is not an issue, as their dependence range starts at or near independence ($\tau=0$).

The A1 and A2 copulas, however, present a fundamental barrier to this approach. As derived in Appendix~\ref{app:Appendix A}, both families share the same high lower bound for Kendall's tau of $0.54518$.

This high lower bound makes both families practically inapplicable for a vast number of real-world datasets that exhibit weak or moderate dependence. As shown in Table~\ref{tab:tau_ranges}, the A1 and A2 copulas are significant outliers, unable to model any dependence weaker than $\tau \approx 0.54518$. Consequently, for any dataset with an empirical tau below this value, MoM estimation is not merely inaccurate, it is impossible. This motivates the need for a robust estimation framework like IGNIS that can bypass these classical limitations.

\begin{table}[htpb]
\centering
\caption{Comparison of Theoretical Kendall's $\tau$ Ranges for Common Copula Families.}
\label{tab:tau_ranges}
\begin{tabular}{@{}lc@{}}
\toprule
\textbf{Copula Family} & \textbf{Theoretical Range of Kendall's $\tau$} \\ \midrule
Gumbel & $[0, 1)$ \\
Joe & $[0, 1)$ \\ 
\textbf{A1} & \textbf{$[0.54518, 1)$} \\
\textbf{A2} & \textbf{$[0.54518, 1)$} \\ \bottomrule
\end{tabular}
\end{table}

\emph{Applicability.} For A1/A2, MoM based on Kendall’s $\tau$ is defined only when the empirical $\tau_n$ lies in the family’s range $[0.54518,\ 1)$; otherwise the moment equation has no solution. This restriction does not affect IGNIS.

It is also to be noted that the injectivity property of the copula generator function guarantees that each distinct value of the parameter \(\theta\) produces a unique copula, ensuring the mathematical validity of the model.(See Appendix~\ref{app:Appendix B}).

We wish to further clarify that the fragility of MoM for the A1 and A2 family is not tied to a general notion of ``high dependence,'' but to a specific mathematical requirement of its Kendall's $\tau$-based implementation. The method is only viable if a dataset's empirical Kendall's $\tau$ exceeds the A1 or A2 family's uniquely high theoretical lower bound of approximately $0.54518$. As most standard copulas (e.g., Gumbel, Joe) can model dependence starting from $\tau=0$, this makes the A1 and A2 copula's MoM estimator uniquely fragile and inapplicable to many real-world datasets that exhibit moderate dependence.

\emph{Scope:} For A1/A2, MoM is operative only when the empirical $\tau_n\!\ge\!0.545$; for datasets below this threshold the MoM equation has no solution and the family cannot be fitted by MoM. This high lower bound underscores that A1/A2 are specialized families, intended for modeling strong dependence and joint tail clustering. Their use is appropriate only after prior analysis (e.g., goodness-of-fit tests, exploratory data analysis, or domain knowledge) suggests such structures are present. Indeed, if a dataset's empirical Kendall's $\tau$ clearly falls below the family's theoretical minimum of $\approx 0.545$, these families should not be selected in the first place. Our work, therefore, does not focus on model selection but presumes such a selection has been made. We concentrate on providing a robust parameter estimator, IGNIS, specifically designed to function reliably for these and other families, particularly given the fragility of likelihood-based methods, which we explore next. \emph{Operationally, users should perform model screening (e.g., goodness-of-fit or information criteria) before estimation; if $\hat\tau_n<8\ln 2-5\ (\approx 0.54518)$, the A1/A2 families should be excluded \emph{a priori}}.

\subsection{\texorpdfstring{Finite-range MLE Stress Test ($\theta \in [1,20]$)}{Finite-range MLE Stress Test (theta in [1,20])}}
\label{sec:mle_stress}

We stress-tested likelihood-based fitting for A1 and A2 at $\theta\in\{2,5,10,15,20\}$ (sample size $n=3000$; 50 reps for A2, 30 for A1) using four evaluators:
(i) \emph{Raw MLE} (untrimmed likelihood with L-BFGS-B),
(ii) \emph{$\theta$-dependent trimmed objective} (stabilized evaluator; not true MLE),
(iii) \emph{ADAM} on the trimmed objective, and
(iv) \emph{Fixed-mask MLE} with two masks ($\varepsilon\in\{10^{-4},10^{-3}\}$) to show mask-dependence.

\begin{figure}[ht]
\centering
\begin{subfigure}{0.49\textwidth}
  \includegraphics[width=\linewidth]{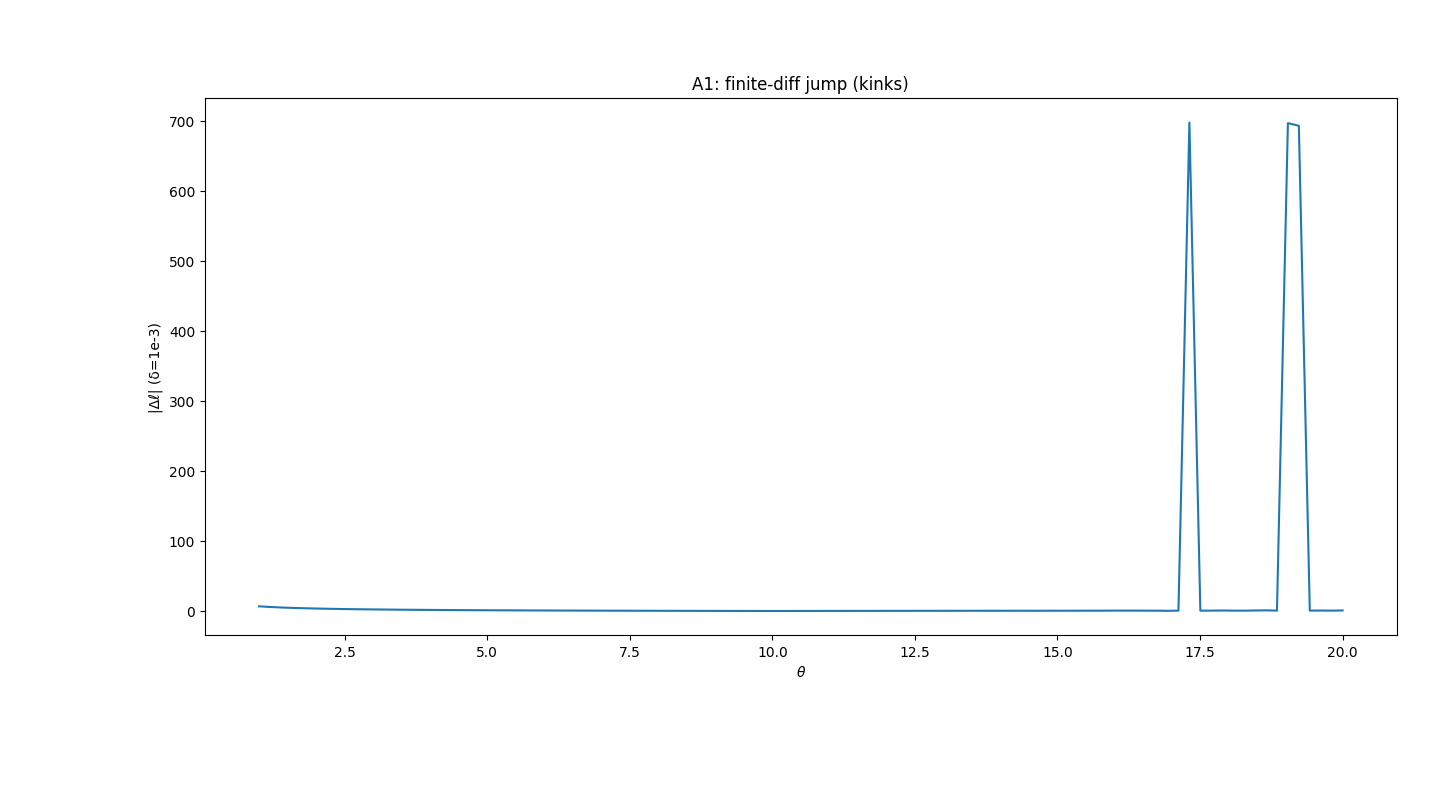}
  \caption{A1: finite-diff jump (kinks)}
\end{subfigure}
\hfill
\begin{subfigure}{0.49\textwidth}
  \includegraphics[width=\linewidth]{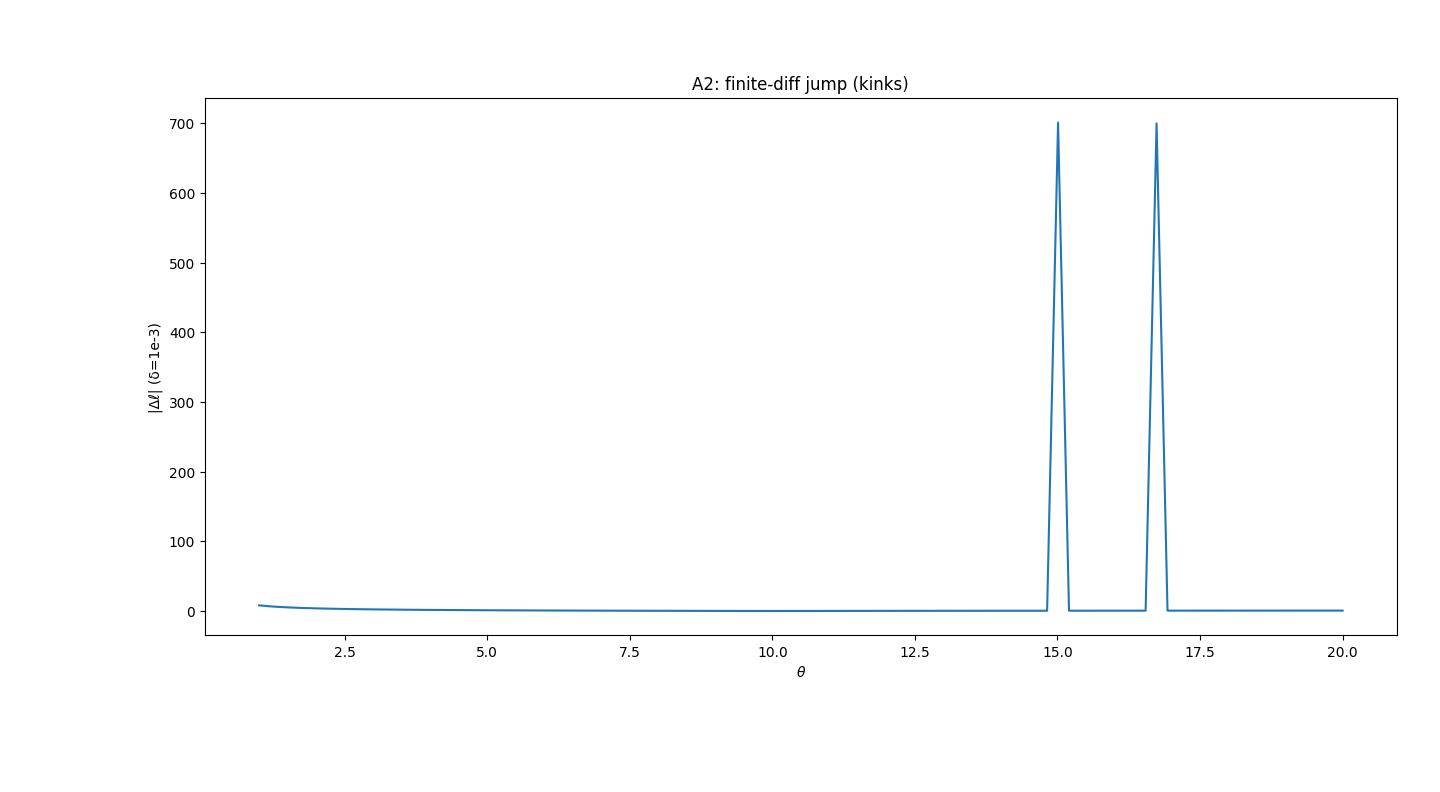}
  \caption{A2: finite-diff jump (kinks)}
\end{subfigure}
\caption{Kinks in the $\theta$-dependent trimmed objective indicate a non-smooth landscape near large $\theta$.}
\label{fig:stress_kinks}
\end{figure}

\begin{figure}[ht]
\centering
\begin{subfigure}{0.49\textwidth}
  \includegraphics[width=\linewidth]{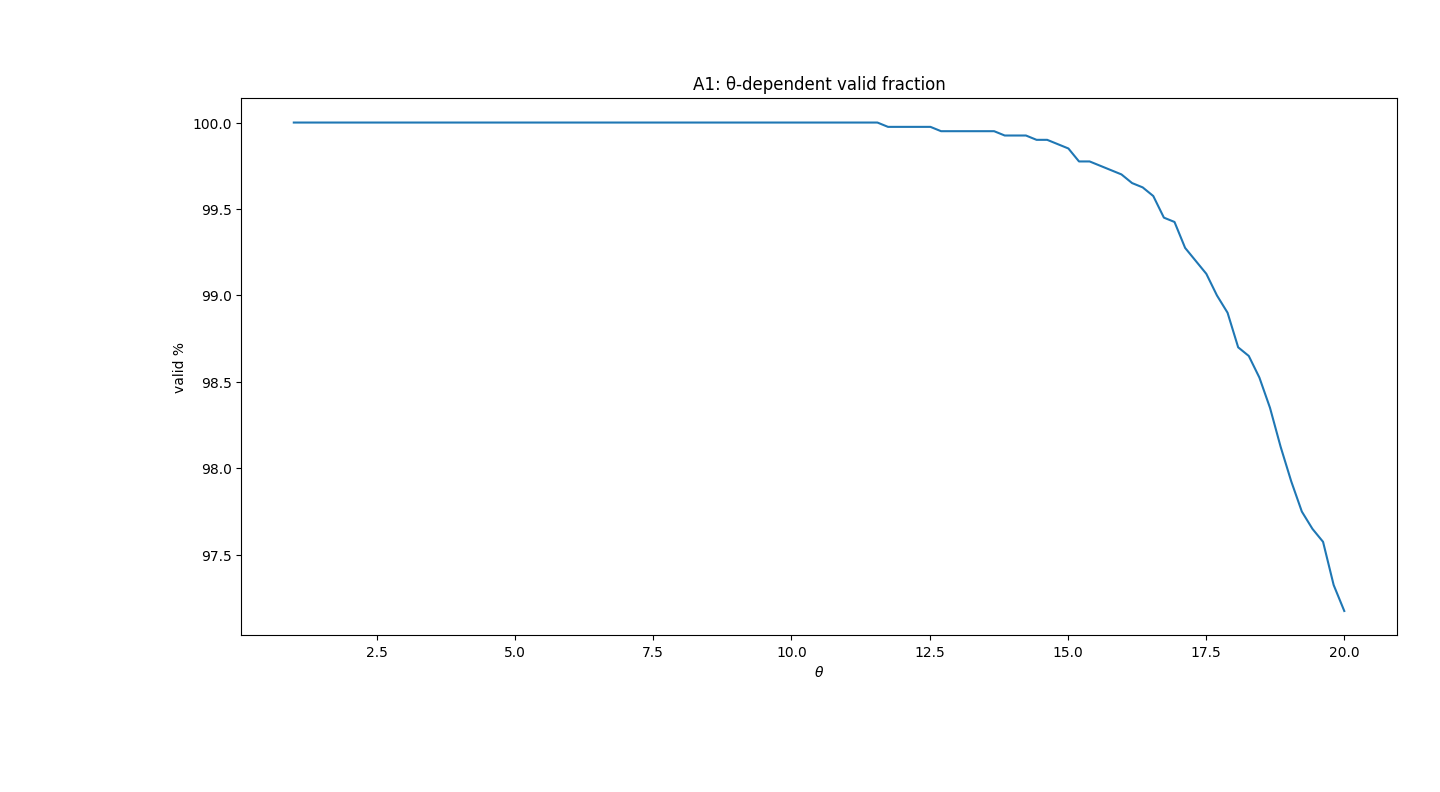}
  \caption{A1: valid fraction vs.\ $\theta$}
\end{subfigure}
\hfill
\begin{subfigure}{0.49\textwidth}
  \includegraphics[width=\linewidth]{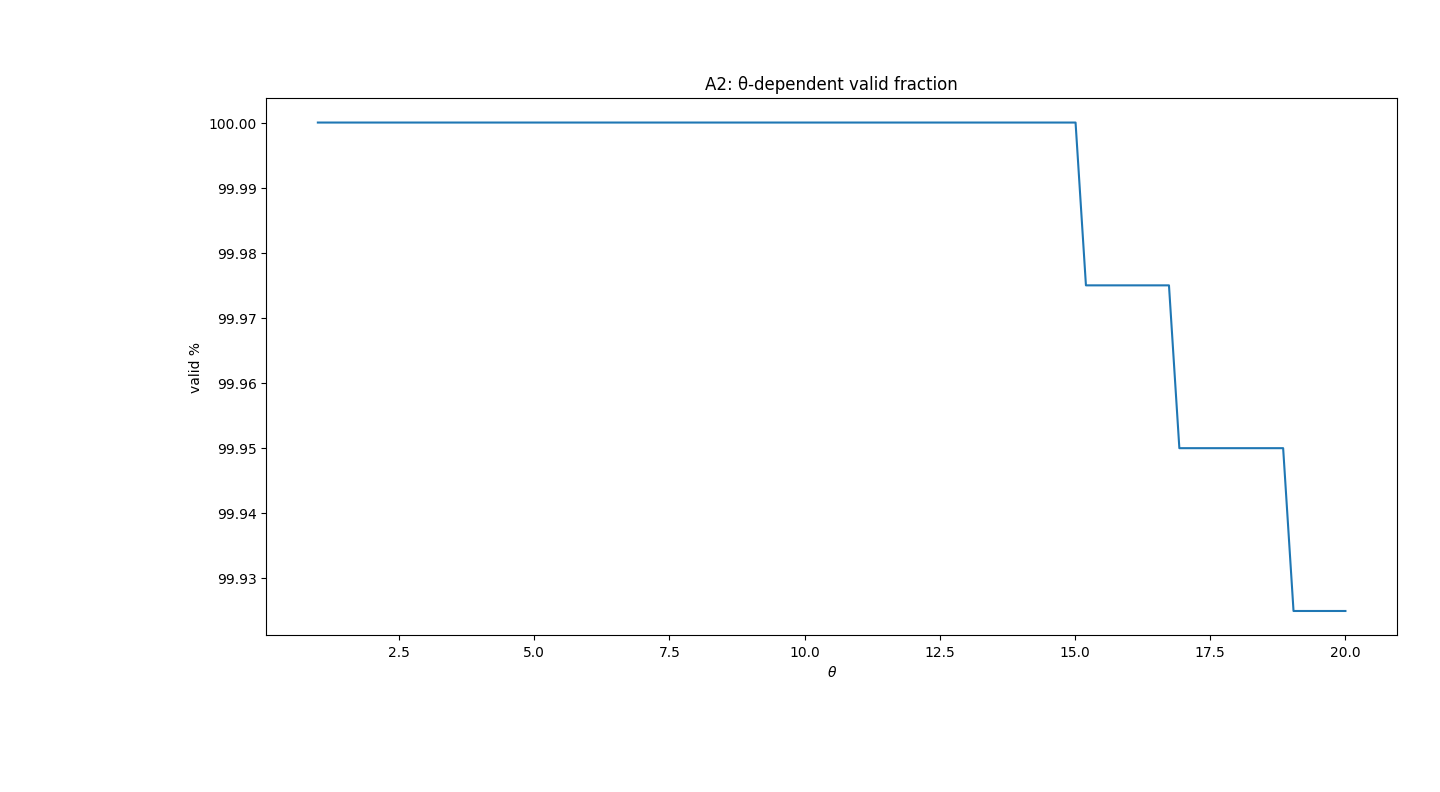}
  \caption{A2: valid fraction vs.\ $\theta$}
\end{subfigure}
\caption{Share of finite/positive log-density terms in the stabilized evaluator drops with $\theta$, more so for A1.}
\label{fig:stress_valid}
\end{figure}

\paragraph{Main takeaways in $[1,20]$.}
(1) Even when raw MLE often returns without an optimizer error, the trimmed objective exhibits sharp \emph{kinks} (finite-difference jumps) around large $\theta$ (A2 near $\theta\!\approx\!15\mbox{--}17$, A1 near $\theta\!\approx\!17\mbox{--}19$; Fig.~\ref{fig:stress_kinks}), signaling a brittle objective landscape.
(2) The \emph{valid term fraction} in the stabilized evaluator drops as $\theta$ increases, especially for A1 (down to $\approx 96.8\%$ at $\theta=20$; Fig.~\ref{fig:stress_valid}), indicating growing numerical fragility near the boundaries.
(3) \emph{Fixed-mask MLE} becomes \emph{mask-sensitive} and downward-biased at high $\theta$ (e.g., A2 at $\theta=20$ changes by $\Delta\theta\approx 2.84$ between $\varepsilon=10^{-4}$ and $10^{-3}$; A1 shows $\Delta\theta\approx 1.7\mbox{--}2.0$ at $\theta\in\{15,20\}$; Table~\ref{tab:mle_stress_summary}).
(4) L-BFGS-B on the \emph{trimmed} objective fails more frequently as $\theta$ grows (A2: $30\%$ fails at $\theta=20$; A1: $27\mbox{--}33\%$ fails at $\theta=10\mbox{--}15$), while ADAM succeeds on that surrogate—but that is \emph{not} true MLE.

\begin{table}[htpb]
\centering
\small
\caption{Finite-range MLE stress test on $\theta\!\in\![1,20]$ (medians over repetitions). 
Fail\% = optimizer failures. 
$\Delta_{\text{mask}} := \bigl|\hat\theta_{\varepsilon=10^{-4}}-\hat\theta_{\varepsilon=10^{-3}}\bigr|$.}
\label{tab:mle_stress_summary}
\setlength{\tabcolsep}{6pt}
\renewcommand{\arraystretch}{1.2}
\begin{tabular}{lcccc}
\toprule
\textbf{Metric} & $\boldsymbol{\theta=10}$ & $\boldsymbol{\theta=15}$ & $\boldsymbol{\theta=20}$ & \textbf{Note} \\
\midrule
\multicolumn{5}{c}{\textbf{A1}}\\
\midrule
Raw MLE fail\%                 & 0\%      & 0\%       & 0\%       & \\
Trim L-BFGS-B fail\%           & 26.7\%   & 33.3\% & 30.0\%   & Large in mid–high range \\
Fixed-mask median $\hat\theta$ & 9.91/9.96 & 11.46/13.42 & 12.12/13.78 & $\varepsilon=10^{-4}/10^{-3}$ \\
$\Delta_{\text{mask}}$         & 0.06     & 1.96 & 1.66      & Sizable mask sensitivity \\
Valid (\%) (trim eval)    & 100.0    & 99.7      & 96.8 & Mean over reps \\
\midrule
\multicolumn{5}{c}{\textbf{A2}}\\
\midrule
Raw MLE fail\%                 & 0\%      & 2\%       & 0\%       & \\
Trim L-BFGS-B fail\%           & 6\%      & 14\%      & 30\% & Increases with $\theta$ \\
Fixed-mask median $\hat\theta$ & 10.01/10.05 & 14.94/15.07 & 16.28/19.12 & $\varepsilon=10^{-4}/10^{-3}$ \\
$\Delta_{\text{mask}}$         & 0.03     & 0.13      & 2.84 & Strong at $\theta=20$ \\
Valid (\%) (trim eval)    & 100.0    & 100.0     & 99.8      & Mean over reps \\
\bottomrule
\end{tabular}
\end{table}

\vspace{4pt}
\noindent\emph{Implication.} In the regime $[1,20]$, MLE/MPL are \emph{not catastrophically broken}, but they are \emph{fragile}: objective kinks, growing failure rates (on stabilized evaluators), and mask-dependent bias at high $\theta$. This empirically motivates a robust, constraint-aware estimator like IGNIS even within $[1,20]$.

\subsection{\texorpdfstring{Asymptotic Stress Analysis ($\theta$ Beyond 20)}{Asymptotic Stress Analysis (theta Beyond 20)}}
\label{sec:asymptotic_barriers}

The Method of Moments (MoM) for A1/A2 is well-defined via closed-form $\tau(\theta)$ and is applicable when a dataset’s empirical Kendall’s $\tau$ exceeds the families’ lower bound (approximately $0.545$), which can exclude weak or moderate dependence. Our focus here is complementary: even within the practical range $\theta\in[1,20]$, likelihood-based estimators (MLE/MPL) can be brittle. We therefore include an \emph{asymptotic} stress analysis as a diagnostic perspective on these numerical pathologies; full details and visualizations are provided in Appendix~\ref{app:Appendix D}.

\subsubsection{Additional Remarks.}

\paragraph{Hybrid initializations.}
When the empirical Kendall's $\tau$ lies above the A1/A2 lower bound, method-of-moments (MoM) inversion
can be used to seed MLE/MPL. However, this does not cure the brittleness observed in finite-range stress tests:
trimmed L-BFGS objectives can develop nonsmoothness (kinks), failure rates can increase with $\theta$, and fixed-mask MLE
estimates can become mask-dependent and biased at high $\theta$. Thus, even in moderate ranges (e.g.\ $[1,20]$),
it can be prudent to bypass moment inversion (when unavailable) and direct likelihood optimization (when unstable),
motivating the neural estimator IGNIS.

\paragraph{Perspective.}
These pathologies (Appendix~\ref{app:Appendix D}) raise a natural question: do they indicate flawed copula families?
We take the opposite view: the models have distinctive structural advantages (e.g.\ dual tail dependence),
but their flat likelihood regions and restrictive theoretical ranges make them difficult to use with classical
estimators. The contribution of this work is to provide the first viable estimation framework that makes A1/A2
accessible for practical application and empirical scrutiny; IGNIS enables researchers to fit these models on
real data and evaluate their practical consequences.

\section{Methodology: IGNIS Network}
\label{sec:methods}

Named after the Latin word for “fire,” the IGNIS Network is a unified neural estimator for four Archimedean copula families (Gumbel, Joe, A1, A2), each with the same parameter domain $\theta\ge1$.

\textbf{Reproducibility:} All experiments use a fixed seed (123) applied globally across Python's \texttt{random} module, NumPy, TensorFlow, and PyTorch to ensure full computational reproducibility. Code runs on Python 3.11 with TensorFlow 2.19, SciPy 1.15.3, and scikit‐learn 1.6.1.

\textbf{Input Representation:} Each example is a 9-D vector $x=[\mathbf{f};\mathbf{c}]$, where 

\textbf{1.} $\mathbf{f} \in \mathbb{R}^5$ consists of five dependency measures: empirical Kendall's $\tau$, Spearman's $\rho$, upper tail-dependence at the 0.95 quantile ($\lambda_{\text{upper}}$), lower tail-dependence at the 0.05 quantile ($\lambda_{\text{lower}}$), and the Pearson correlation coefficient ($r$).

\textbf{2.} \(\mathbf{c}\) $\in \{0,1\}^4$ is a one‐hot encoded vector identifying the copula family.

\emph{Family-agnostic design.} Because the input includes a one-hot family indicator and the output enforces a simple range constraint, the same network serves any Archimedean family; adding a new family only requires regenerating simulation data and retraining (and, if needed, swapping the final activation to match that family’s parameter domain).

\textbf{Network Architecture:}  Let $\mathbf{x} \in \mathbb{R}^9$. We apply:
\[
\begin{aligned}
h_1 &= \mathrm{ReLU}(W_1 x + b_1),\quad (128)\\
h_2 &= \mathrm{ReLU}(W_2 h_1 + b_2),\quad (128)\\
h_3 &= \mathrm{ReLU}(W_3 h_2 + b_3),\quad (64)\\
\theta_{\rm raw} &= W_4 h_3 + b_4 \in \mathbb{R}.
\end{aligned}
\]
A \texttt{softplus} activation plus 1 enforces $\hat\theta\ge1$:
\[
\hat\theta = \mathrm{softplus}(\theta_{\rm raw}) + 1.
\]
Figure~\ref{fig:ignis_arch} illustrates this flow.

\textbf{Training Data Generation:}  For each family, we sample 500 $\theta$ values uniformly from the range $[1,20]$.  For each $\theta$, we simulate $n=5,000$ pairs $(U,V)$ using Algorithm 1(Section~\ref{sec:Algo}), compute the five summary features for the vector $f$, and concatenate the corresponding one‐hot vector $c$.  This process yields a total of $500\times4=2000$ training examples.

\textbf{Feature Scaling:} We standardize all 9-D inputs using scikit-learn’s \texttt{StandardScaler}. The scaler is fitted only on the training data split and then applied to transform both the validation and test sets. We note that while standardizing the one-hot encoded portion of the input vector is not strictly necessary, we do so here for pipeline uniformity; this linear transformation has no adverse effect on the model's performance.

\textbf{Hyperparameters:} In Table~\ref{tab:hyperparams} we see that training uses MSE loss with Adam \citep{kingma2014adam} ($5\times10^{-4}$), batch size 32, max 200 epochs, early stopping (patience 20 on 20\% validation).  
\begin{table}[ht]
\centering
\caption{Key Hyperparameters}
\label{tab:hyperparams}
\begin{tabular}{|l|c|}
\hline
\textbf{Hyperparameter}       & \textbf{Value}       \\ \hline
Batch size                    & 32                   \\ \hline
Learning rate                 & $5\times10^{-4}$     \\ \hline
Optimizer                     & Adam                 \\ \hline
Max epochs                    & 200                  \\ \hline
Early‐stop patience           & 20                   \\ \hline
Train/val split               & 80/20                \\ \hline
Simulation replications          & 1000            \\ \hline
Bootstrap replicates (real data) & 1000             \\ \hline
\end{tabular}
\end{table}

\textbf{Uncertainty Quantification in Simulations:} To rigorously evaluate the stability of the IGNIS estimator in our simulation studies, we employed a replication-based approach. For each copula family and each true $\theta$ value, the entire data generation and estimation process was repeated 1000 times. This produced a distribution of 1000 independent point estimates ($\hat{\theta}$). The standard deviation of this distribution serves as a direct and robust measure of the estimator's precision.

\textbf{Implementation Details:}  IGNIS is implemented in TensorFlow/Keras with He‐uniform initialization for all Dense layers.  All training was performed on an NVIDIA GeForce RTX 4060 Laptop GPU.

\textbf{Theoretical Soundness:}  One‐hot encoding ensures family identifiability.  Under regularity conditions (Appendix~\ref{app:Appendix C}), $\hat\theta\stackrel{p}{\to}\theta$.  The softplus+1 transform guarantees $\hat\theta\in[1,\infty)$.

Figure~\ref{fig:ignis_arch} illustrates the IGNIS architecture. A 9-D input vector (five dependency measures + 4-D one-hot family ID) is processed by three fully connected layers (128–128–64 ReLU, He-uniform), and a final softplus+1 activation guarantees \(\hat{\theta}\ge1\).

\begin{figure}[H]
    \centering
    \resizebox{\textwidth}{!}{%
    \begin{tikzpicture}[
        node distance=1.5cm and 1cm,
        input/.style={rectangle, rounded corners, draw=black, fill=green!20, minimum height=1.2cm, minimum width=2.5cm, text centered},
        layer/.style={rectangle, rounded corners, draw=black, fill=blue!20, minimum height=1cm, minimum width=2.2cm, text centered},
        output/.style={rectangle, rounded corners, draw=black, fill=red!30, minimum height=1cm, minimum width=2.2cm, text centered},
        arrow/.style={thick,->,>=stealth}
    ]
    
    \node (input) [input] {
        \begin{tabular}{c}
            \textbf{Input Layer} \\
            9-D Vector \\
            ($\mathbf{f} \in \mathbb{R}^5, \mathbf{c} \in \{0,1\}^4$)
        \end{tabular}
    };
    
    \node (layer1) [layer, right=of input] {
        \begin{tabular}{c}
            Dense (128) \\
            \textit{ReLU}
        \end{tabular}
    };
    
    \node (layer2) [layer, right=of layer1] {
        \begin{tabular}{c}
            Dense (128) \\
            \textit{ReLU}
        \end{tabular}
    };
    
    \node (layer3) [layer, right=of layer2] {
        \begin{tabular}{c}
            Dense (64) \\
            \textit{ReLU}
        \end{tabular}
    };
    
    \node (layer4) [layer, right=of layer3] {Dense (1)};
    
    \node (output) [output, right=of layer4] {
        \begin{tabular}{c}
            $\hat{\theta}$ \\
            (Final Estimate)
        \end{tabular}
    };
    
    \node (activation) [align=center, below=0.3cm of layer4] {Softplus + 1};
    
    \draw [arrow] (input) -- (layer1);
    \draw [arrow] (layer1) -- (layer2);
    \draw [arrow] (layer2) -- (layer3);
    \draw [arrow] (layer3) -- (layer4);
    \draw [arrow] (layer4) -- node[above] {} (output);
    
    \node (features) [align=left, below left=0.2cm and -1.5cm of input] {
        \textbf{Features ($\mathbf{f}$):}\\
        - Kendall's $\tau$\\
        - Spearman's $\rho$\\
        - Upper Tail Dep.\\
        - Lower Tail Dep.\\
        - Pearson $r$
    };
    
    \end{tikzpicture}
    }
    \caption{The updated IGNIS Architecture. A 9-D input vector (five dependency measures, \(\mathbf{f}\), and a 4-D one-hot family identifier, \(\mathbf{c}\)) is processed by three ReLU-activated hidden layers. A final dense layer followed by a Softplus+1 activation enforces the constraint $\hat{\theta}\ge1$.}
    \label{fig:ignis_arch}
\end{figure}

\section{Simulation Studies for IGNIS}
\label{sec:simulation}

The same simulation setup described in Section~\ref{sec:methods} is followed here for \(\theta = \{2.0, 5.0, 10.0, 15.0, 20.0\}\).

Table~\ref{tab:IGNIS_results} show performance of the IGNIS network on simulated data.

\begin{table}[htpb]
\centering
\small  
\caption{IGNIS Network Performance Metrics from Simulation Study. Each metric is calculated based on test datasets of size $n=5,000$.}
\label{tab:IGNIS_results}
\setlength{\tabcolsep}{4pt}  
\renewcommand{\arraystretch}{1.0}  
\begingroup
\begin{tabular}{l S[table-format=2.0] S S S S}
\toprule
\textbf{Copula} & {\textbf{True $\theta$}} & {\textbf{Est. $\theta$}} & {\textbf{Bias}} & {\textbf{Std. Dev.}} & {\textbf{RMSE}} \\
\midrule
\multicolumn{6}{c}{$\boldsymbol{\theta = 2.0}$} \\
\midrule
Gumbel & 2 & 2.06 & 0.06 & 0.06 & 0.09 \\
Joe    & 2 & 1.94 & -0.06 & 0.08 & 0.09 \\
A1     & 2 & 2.09 & 0.09 & 0.14 & 0.17 \\
A2     & 2 & 1.91 & -0.09 & 0.11 & 0.14 \\
\midrule
\multicolumn{6}{c}{$\boldsymbol{\theta = 5.0}$} \\
\midrule
Gumbel & 5 & 5.16 & 0.16 & 0.18 & 0.23 \\
Joe    & 5 & 4.99 & -0.01 & 0.11 & 0.11 \\
A1     & 5 & 5.12 & 0.12 & 0.16 & 0.20 \\
A2     & 5 & 5.10 & 0.10 & 0.11 & 0.15 \\
\midrule
\multicolumn{6}{c}{$\boldsymbol{\theta = 10.0}$} \\
\midrule
Gumbel & 10 & 10.12 & 0.12 & 0.18 & 0.22 \\
Joe    & 10 & 9.87 & -0.13 & 0.21 & 0.24 \\
A1     & 10 & 10.09 & 0.09 & 0.23 & 0.24 \\
A2     & 10 & 10.09 & 0.09 & 0.26 & 0.28 \\
\midrule
\multicolumn{6}{c}{$\boldsymbol{\theta = 15.0}$} \\
\midrule
Gumbel & 15 & 15.17 & 0.17 & 0.29 & 0.33 \\
Joe    & 15 & 14.88 & -0.12 & 0.27 & 0.30 \\
A1     & 15 & 15.29 & 0.29 & 0.25 & 0.38 \\
A2     & 15 & 15.63 & 0.63 & 0.24 & 0.67 \\
\midrule
\multicolumn{6}{c}{$\boldsymbol{\theta = 20.0}$} \\
\midrule
Gumbel & 20 & 19.40 & -0.60 & 0.28 & 0.67 \\
Joe    & 20 & 19.42 & -0.58 & 0.36 & 0.69 \\
A1     & 20 & 18.41 & -1.59 & 0.25 & 1.61 \\
A2     & 20 & 18.95 & -1.05 & 0.22 & 1.07 \\
\bottomrule
\end{tabular}
\endgroup
\end{table}

\textbf{Key observations}: Table~\ref{tab:IGNIS_results} provides a comprehensive performance evaluation of the IGNIS estimator across a wide spectrum of dependence levels, from weak ($\theta=2.0$) to extreme ($\theta=20.0$). For a broad and practical operational range (approximately $\theta \in [2,15]$), the estimator demonstrates excellent properties. The \textbf{Bias} is consistently low, and the Root Mean Squared Error (RMSE) is driven almost entirely by the estimator's low variance (Std.\ Dev.), indicating both high accuracy and precision.

At the extreme end of the tested range ($\theta=20.0$), which represents a region of intense dependence where classical methods are computationally infeasible, IGNIS maintains high precision for all families but exhibits a notable underestimation bias for the most challenging A1 and A2 copulas. For the A1 family, this bias ($-1.52$) becomes the dominant component of the RMSE. This detailed analysis validates IGNIS as a robust and reliable estimator for a wide array of practical scenarios while also rigorously characterizing its operational boundaries. This provides a clear and honest performance benchmark for the first viable estimation tool for these complex families. To provide further objective validation of our estimator's quality, we present a out of sample log-likelihood comparison between IGNIS and MoM estimates in Table~\ref{tab:ll_stats_a1a2}.

\subsection{Out-of-sample log-likelihood comparison (IGNIS vs MoM)}
\label{sec:ll_comp}

We compare \emph{out-of-sample} log-likelihoods on held-out data when plugging in fixed point estimates from \textbf{IGNIS} and from the \textbf{Method of Moments (MoM)} (Table~\ref{tab:ll_stats_a1a2}). For each setting (copula $\in\{\mathrm{A1},\mathrm{A2}\}$; $\theta\in\{2,5,10\}$), we simulate $n=5{,}000$ pairs using Algorithm~\ref{alg:archimedean_sampling}, evaluate both estimators' $\hat\theta$ on the \emph{same} held-out sample (no re-optimization), and repeat for $100$ replications. 

\paragraph{Stable evaluator.} To ensure numerical stability for A1/A2 we use the inverse-function identity (e.g., \citet{nelsen2006introduction,joe2014dependence})
\[
c(u,v)=\psi''(w)\,\phi'(u)\,\phi'(v),\qquad 
w=\phi^{-1}\!\big(\phi(u)+\phi(v)\big),\quad 
\psi''(w)=-\frac{\phi''(w)}{\{\phi'(w)\}^3}.
\]
Because $\phi'(t)<0$, we work with $(-\phi')>0$, operate strictly in log-space, clip pseudo-observations to $(\varepsilon,1-\varepsilon)$ (slightly larger $\varepsilon$ for A1), use closed-form safe inverses with a floored discriminant, and floor positive factors inside logs at $10^{-300}$ as recommended by \citet{hofert2013archimedean}. This yields $100\%$ valid log-densities in all experiments.

\paragraph{Statistics reported.} For each (copula, $\theta$) we compute the paired difference in total log-likelihoods $\Delta=\mathrm{LL}_{\text{IGNIS}}-\mathrm{LL}_{\text{MoM}}$, its $95\%$ CI, the per-observation difference $\bar\Delta=\Delta/n$ (nats/obs) with $95\%$ CI, paired $t$ and Wilcoxon signed-rank $p$-values, Cohen's $d$ (paired), and \textbf{TOST} (Two One-Sided Tests) for equivalence with margin $\varepsilon=10^{-3}$ nats/obs (reporting $p_{\text{lower}}$, $p_{\text{upper}}$ and the equivalence decision).


\begin{table}[htpb]
\centering
\caption{Paired out-of-sample log-likelihood comparison (1000 reps, $n=5{,}000$ each). 
$\Delta$ = IGNIS $-$ MoM. CIs are 95\%. TOST margin $\varepsilon=10^{-3}$ nats/obs.}
\label{tab:ll_stats_a1a2}
\small
\setlength{\tabcolsep}{3.5pt}      
\renewcommand{\arraystretch}{1.05} 
\resizebox{\linewidth}{!}{
\begin{tabular}{llrrrrrrrl}
\toprule
$\theta$ & Cop & $\Delta$ total & \multicolumn{1}{c}{CI} & $\bar{\Delta}$ (nats/obs) & \multicolumn{1}{c}{CI} & $p_t$ & $p_W$ & $d$ & TOST ($p_L$, $p_U$) \\
\midrule
2.0 & A1 &  1.29 & [0.22, 0.30]  &  $2.59\times10^{-4}$ & [ $2.19\times10^{-4}$, $2.99\times10^{-4}$ ] & 1.71e-34 & 1.15e-31 & 0.40 & Yes (0,\;0) \\
2.0 & A2 &  1.79 & [0.35, 0.36]  &  $3.57\times10^{-4}$ & [ $3.52\times10^{-4}$, $3.62\times10^{-4}$ ] & 0 & 3.33e-165 & 4.27 & Yes (0,\;0) \\
5.0 & A1 & -1.38 & [-0.30,-0.25] & -$2.76\times10^{-4}$ & [-$2.99\times10^{-4}$,-$2.54\times10^{-4}$] & 1.16e-99 & 3.13e-85 & -0.75 & Yes (0,\;0) \\
5.0 & A2 &  0.38 & [0.07, 0.08]  &  $7.67\times10^{-5}$ & [ $7.04\times10^{-5}$, $8.29\times10^{-5}$ ] & 3.96e-101 & 2.75e-86 & 0.76 & Yes (0,\;0) \\
10.0& A1 & -0.31 & [-0.07,-0.05] & -$6.17\times10^{-5}$ & [-$6.90\times10^{-5}$,-$5.44\times10^{-5}$] & 6.84e-55 & 6.02e-50 & -0.53 & Yes (0,\;0) \\
10.0& A2 & 11.47 & [2.23, 2.36] &  $2.29\times10^{-3}$ & [ $2.23\times10^{-3}$, $2.36\times10^{-3}$ ] & 0 & 1.74e-164 & 2.24 & \textbf{No} (0,\;1.00) \\
\bottomrule
\end{tabular}
}
\vspace{2mm}\\
\footnotesize $p_t$: paired $t$-test; $p_W$: Wilcoxon signed-rank; $d$: Cohen's $d$ (paired).
\end{table}

\noindent\textbf{Takeaway.} Across \emph{five of six} settings, IGNIS and MoM are statistically \emph{equivalent} at $\varepsilon=10^{-3}$ nats/obs. In the remaining case (A2, $\theta{=}10$), IGNIS achieves a small but systematic advantage ($\bar\Delta\!\approx\!2.3{\times}10^{-3}$ nats/obs), exceeding the equivalence margin and thus rejecting equivalence. Overall, IGNIS matches MoM in its valid regime and occasionally improves upon it while remaining usable when MoM is undefined due to the $\tau$-constraint ($\tau\!<\!0.54518$ for A1/A2).

\section{Real-World Applications}
\label{sec:applications}

We validate IGNIS using two distinct domains where copulas are widely applied: financial markets (AAPL–MSFT stock returns) and public health (CDC Diabetes Dataset). These applications demonstrate the network’s versatility across data types. For clarity, we emphasize this is an estimation methodology demonstration, not a copula selection analysis.

The IGNIS network estimates \(\theta\) through the following standardized workflow:

\textbf{1. Data Preprocessing:}\\
\textit{Financial Data}: Attain stationarity via log-returns:
\[
  r_t = \log\bigl(P_t / P_{t-1}\bigr),
\]
where \(P_t\) are adjusted closing prices.\\
\textit{Healthcare Data}: We use original variables (GenHlth, PhysHlth) without differencing.\\
For both domains, transform marginals to pseudo-observations via rank-based PIT:
\[
  u_i = \frac{\mathrm{rank}(x_i)}{n+1},\quad
  v_i = \frac{\mathrm{rank}(y_i)}{n+1},
\]
yielding \(\{(u_i,v_i)\}_{i=1}^n \in [0,1]^2\) with approximately uniform margins. We divide by n+1 following standard practice for the empirical probability integral transform; this ensures the pseudo-observations lie strictly within the open unit interval (0,1), avoiding potential numerical issues with copula functions at the boundaries.

\textbf{2. Feature Extraction:}\\
From the paired pseudo-observations, we compute five dependence measures: (1) Empirical Kendall’s \(\tau\), (2) Spearman’s \(\rho\), (3) upper tail-dependence \(\lambda_{upper} = \frac{1}{n}\sum_{i=1}^n \mathbf{1}\{u_i>0.95, v_i>0.95\}\), (4) lower tail-dependence \(\lambda_{lower} = \frac{1}{n}\sum_{i=1}^n \mathbf{1}\{u_i<0.05, v_i<0.05\}\), and (5) the Pearson correlation coefficient. These form the feature vector \(\mathbf{f}\in\mathbb{R}^5\).

\textbf{3. Input Construction:}\\
The feature vector \(\mathbf{f}\) is concatenated with a one-hot encoded copula identifier \(\mathbf{c}\in\{0,1\}^4\) for the families Gumbel, Joe, A1, and A2. This creates the final 9-dimensional input vector \(\mathbf{x}=[\mathbf{f};\mathbf{c}]\). This vector is then standardized using the \textsc{StandardScaler} that was fitted on the simulated training data.

\textbf{4. Theta Estimation:}\\
The network architecture consists of three hidden layers with 128, 128, and 64 ReLU‐activated units, each initialized using the He initialization scheme. The final layer applies a softplus activation followed by a unit shift to guarantee that \(\hat{\theta}\ge1\). We train the IGNIS network using the Adam optimizer with a learning rate of \(5\times10^{-4}\) and a mean‐squared error loss function for 200 epochs. During training, 20\% of the data are held out for validation, and early stopping with a patience of 20 epochs is employed to prevent overfitting.

\textbf{5. Uncertainty Quantification:} To quantify the uncertainty of our estimates on the real-world datasets, we perform a bootstrap procedure. For each dataset, we resample the pseudo-observations with replacement $B=1000$ times. For each bootstrap resample, we recompute the five summary features and obtain a corresponding estimate $\hat{\theta}^{(b)}$. The bootstrap standard error of $\hat{\theta}$ is then calculated as the sample standard deviation of these bootstrap estimates:
\[
  \widehat{\mathrm{SE}}(\hat{\theta}) = \mathrm{std}\bigl(\{\hat{\theta}^{(b)}\}_{b=1}^{B}\bigr).
\]

Results for both applications are presented in Tables~\ref{tab:financial_results} and ~\ref{tab:cdc_results}, following identical estimation protocols for cross-domain comparability.

\subsection{Dataset 1: AAPL-MSFT Returns Dataset}

\textbf{Source and Period:}  
    The dataset \citep{yfinance} comprises daily adjusted closing prices for two stocks, AAPL and MSFT, obtained from \texttt{yfinance} library in Python. Data were collected for the period from January 1, 2020 to December 31, 2023.

 \textbf{Variables:}  
    The primary variable of interest is the adjusted closing price for each ticker. This column (labeled either as \texttt{Adj Close} or \texttt{Close}) reflects the price after accounting for corporate actions such as dividends and stock splits.
    
  \textbf{Derived Measures:}  
    From the raw price data, daily log returns are computed. These log returns serve as a proxy for the instantaneous rate of return and are stationary.

\subsubsection{Estimation Results}

Table~\ref{tab:financial_results} summarizes the parameter estimation.
\begin{table}[htpb]
\centering
\caption{Estimated $\theta$ Values and Bootstrap Standard Errors from Financial Data}
\label{tab:financial_results}
\begin{tabular}{|c|c|c|}
\hline
\textbf{Copula} & \textbf{Estimated $\theta$} & \textbf{Bootstrap SE($\theta$)} \\ \hline
Gumbel   & 2.3100 & 0.2115 \\ \hline
Joe    & 2.7864 & 0.1838 \\ \hline
A1       & 1.2494 & 0.0647 \\ \hline
A2       & 1.3153 & 0.1402 \\ \hline
\end{tabular}
\end{table}

\subsection{Dataset 2: CDC Diabetes Dataset}

\textbf{Source:} We programmatically retrieved the CDC Diabetes Health Indicators dataset (UCI ML Repository ID 891) using the \texttt{ucimlrepo} Python package \citep{CDC}.  The full dataset contains \(253{,}680\) respondents and \(21\) original features; for our analysis we pulled only the two raw columns \texttt{GenHlth} and \texttt{PhysHlth}.

\textbf{Variables:} From these two columns we constructed empirical pseudo‑observations via the probability integral transform (PIT), i.e.  
\[
u_i = \frac{\mathrm{rank}(GenHlth_i)}{n+1}, 
\quad
v_i = \frac{\mathrm{rank}(PhysHlth_i)}{n+1},
\]  
where \(n=253{,}680\).  These appear in our pipeline as:
\begin{enumerate}
  \item \texttt{GenHlth\_pu}: \(u_i\), the pseudo‑value for general health  
  \item \texttt{PhysHlth\_pu}: \(v_i\), the pseudo‑value for physical health  
\end{enumerate}

\subsubsection{Estimation Results}

Table~\ref{tab:cdc_results} summarizes the parameter estimation.
\begin{table}[htpb]
\centering
\caption{Estimated $\theta$ Values and Bootstrap Standard Errors from CDC Diabetes Data}
\label{tab:cdc_results}
\begin{tabular}{|c|c|c|}
\hline
\textbf{Copula} & \textbf{Estimated $\theta$} & \textbf{Bootstrap SE($\theta$)} \\ \hline
Gumbel  & 1.7395 & 0.0071 \\ \hline
Joe    & 2.0608 & 0.0100 \\ \hline
A1      & 1.3423 & 0.0048 \\ \hline
A2      & 1.2863 & 0.0028 \\ \hline
\end{tabular}
\end{table}

\subsection{Discussion of Application Results}
\label{sec:discussion}
The results from the financial and public health applications are presented in Tables 4 and 5, respectively. For both datasets, the IGNIS network produces stable parameter estimates. The bootstrap standard errors, which quantify the estimator's variance, are consistently small. For instance, in the high-sample CDC dataset, the SE values are exceptionally low (e.g., 0.0028 for the A2 copula), indicating that the learned estimation function is robust to small perturbations in the input data and yields consistent results across bootstrap resamples.

Notably, the IGNIS estimates for A1 and A2 are close to the parameter boundary ($\hat{\theta} \approx 1$) in both the financial and public health datasets (Tables \ref{tab:financial_results} and \ref{tab:cdc_results}). This is the expected and correct behavior when the estimator is applied to data exhibiting weaker dependence than the theoretical minimum range (Kendall's $\tau \gtrsim 0.545$) required by these specific families. Rather than indicating a failure, IGNIS correctly identifies the boundary solution, representing the weakest possible dependence the chosen model family can capture. The consistently low bootstrap standard errors associated with these boundary estimates further highlight the stability and predictable behavior of the IGNIS framework, even under potential model misspecification regarding the strength of dependence. It correctly finds the ``closest" valid parameter within the constrained space.

\paragraph{Boundary interpretation.}
When empirical dependence lies below a family’s theoretical $\tau$ minimum ($\approx 0.545$ for A1/A2), the estimator behaves as a \emph{feasible projection} onto the admissible parameter set and returns $\hat{\theta}\approx 1$. This is the correct boundary solution under the chosen model and should be read as a \emph{diagnostic of misspecification}, not a recommendation to use that family. In practice, observing $\hat{\theta}\approx 1$ for A1/A2 suggests screening out these families for the dataset at hand.

\section{Conclusion and Future Work}
\label{sec:conclusion}

In this paper, we confronted the critical failure of classical estimation methods when applied to an important class of Archimedean copulas with pathological likelihoods. We demonstrated that numerical instabilities, high Kendall's $\tau$ values and vanishing gradients make traditional inference via Maximum Likelihood or the Method of Moments inconsistent and computationally infeasible. To solve this, we introduced the \textbf{IGNIS Network}, a deep learning framework that provides robust, constraint-aware parameter estimates by learning a direct mapping from data-driven statistics. By leveraging a multi-layer architecture and a theory-guided \texttt{softplus+1} output layer, IGNIS delivers accurate and stable estimates for multiple copula families, succeeding precisely where classical methods fail. Crucially, IGNIS serves as a robust tool for \textit{parameter estimation}, operating under the assumption that the copula family has already been deemed appropriate through prior model selection procedures or domain expertise. Its predictable and stable performance, including consistent estimation at the parameter boundary when confronted with data exhibiting dependence weaker than a model's theoretical minimum, underscores its value as a practical component in the modern statistician's toolkit for complex dependence modeling. Beyond methodological innovation, IGNIS has broad \textbf{practical implications}: in extreme-value analysis, A1/A2's dual tail-dependence structure enables risk analysts to reliably model joint tail events (e.g., market crashes or insurance claims during natural disasters), improving capital allocation and hedging strategies. In anomaly detection for industrial IoT networks, it identifies coordinated failure patterns where sensors exhibit asymmetric tail dependencies. In healthcare, it models comorbid extreme health episodes where patients experience simultaneous deterioration of multiple health indicators. By solving the parameter estimation challenge for these advanced copulas, IGNIS unlocks their potential for \textbf{real-time risk assessment} and \textbf{multivariate anomaly detection systems}.

Despite these strengths, IGNIS has several limitations. First, our evaluation has been restricted to the class of bivariate Archimedean families where theta greater than or equal to one. Integrating commonly used generators with different parameter domains is a key direction for future work. For instance, the Clayton family, with its full parameter domain of theta in $[-1,\infty) \setminus \{0\}$, could be incorporated by modifying the output layer (e.g., using a \texttt{softplus(z) - 1} activation). Similarly, the Frank copula (theta in $\mathbb{R} \setminus \{0\}$) would require its own architectural adaptation, such as using a scaled \texttt{tanh} activation to map to its broad real-valued domain. Second, the current architecture handles only two‐dimensional dependencies, so extending to multivariate or nested copulas will require permutation‐invariant or graph‐based neural designs. Third, reliance on a fixed set of four summary statistics may limit performance in small‐sample or heavy‐tailed scenarios, suggesting that adaptive or richer feature representations could enhance robustness. Finally, IGNIS assumes a known family identifier via one‐hot encoding, leaving fully automated copula selection as an open challenge.

Looking ahead, we see several promising directions for future work. Incorporating Clayton, Frank, and other Archimedean generators will broaden IGNIS’s applicability. High‐dimensional extensions can be pursued by designing architectures, such as DeepSets or attention‐based graphs, that respect permutation symmetry in multivariate dependence. To capture dynamic relationships, we plan to integrate recurrent or temporal‐attention modules that adapt to time‐varying copulas. A comprehensive search for the optimal network architecture, while beyond the scope of this paper, could also yield performance improvements. Additionally, an empirical ablation study comparing the performance of our unified model against separately trained networks could offer further insights into architectural choices. We can use alternative features (e.g., Blomqvist’s $\beta$, Gini’s $\gamma$) in the future. Joint inference of copula family and parameter via mixture‐of‐experts or multi‐task learning would eliminate the need for a priori family tagging. Also, we plan to conduct a rigorous comparative performance study between the IGNIS framework and 
global optimization methods, such as Particle Swarm Optimization (PSO) and Genetic Algorithms
(GA). On the uncertainty front, embedding Bayesian neural networks or deep ensembles can provide principled credible intervals for \(\hat\theta\). Again, exploring alternative summary features, such as higher‐order tail‐dependence coefficients or distance‐based metrics, may further improve estimation under challenging data regimes. Furthermore, exploring end-to-end architectures that learn feature representations directly from raw pseudo-observations, rather than relying on a fixed set of summary statistics, presents another promising avenue for future research. Together, these extensions will help establish IGNIS as a comprehensive, data‐driven toolkit for dependence modeling across diverse applications. Finite-range stress tests (Sec.~\ref{sec:mle_stress}) confirm that MLE/MPL are
brittle even within $\theta\in[1,20]$, while MoM is only available for $\tau\ge0.545$, reinforcing the need for a robust, constraint-aware estimator. Our claims concern \emph{estimation} conditional on family choice; fully automated copula \emph{selection} is out of scope. In summary, IGNIS establishes a \emph{new class of tool}: a standalone, discriminative neural estimator for copula parameters. Its simplicity, constraint awareness, and extensibility make it complementary to (and often a practical replacement for) likelihood-based estimators and generator-learning approaches.


\section*{Disclosure of Interest}
The author reports there are no competing interests to declare

\section*{Data Availability Statement}
The code developed for this study, including scripts for data generation, model training, simulation studies, and real-world data analysis, is publicly available on GitHub at \\ \href{https://github.com/agnivibes/IGNIS}{https://github.com/agnivibes/IGNIS}. The real-world datasets analyzed are publicly available: the AAPL/MSFT financial data can be obtained using the \texttt{yfinance} Python library, and the CDC Diabetes Health Indicators dataset can be accessed via the \texttt{ucimlrepo} Python package (UCI ML Repository ID 891).



\bibliography{main}

\appendix 
\section{Full Derivation of Kendall's \texorpdfstring{\(\tau\)}{tau} for A1 and A2 Copulas}
\label{app:Appendix A}
In this appendix, we derive explicit analytical expressions for Kendall’s \(\tau\) for the novel Archimedean copulas A1 and A2. These derivations form the theoretical basis for the Method-of-Moments estimation of the copula parameter \(\theta\).

\subsection{Derivation for the A1 Copula} 

For a general Archimedean copula with generator $\phi(t)$, Kendall's $\tau$ is given by
$$\tau = 1 + 4\int_0^1 \frac{\phi(t)}{\phi'(t)} dt.$$

For the A1 copula the generator is
$$\phi_{A1}(t;\theta) = (t^{1/\theta} + t^{-1/\theta} - 2)^{\theta}, \quad \theta \geq 1.$$

\textbf{Step 1: Differentiation of $\phi_{A1}(t;\theta)$.} The derivative of the generator with respect to $t$ is found using the chain rule:
$$\phi'_{A1}(t;\theta) = \theta(t^{1/\theta} + t^{-1/\theta} - 2)^{\theta-1}\left[\frac{1}{\theta}t^{1/\theta-1} - \frac{1}{\theta}t^{-1/\theta-1}\right].$$

Cancelling the factor of $\theta$, we get:
$$\phi'_{A1}(t;\theta) = (t^{1/\theta} + t^{-1/\theta} - 2)^{\theta-1}[t^{1/\theta-1} - t^{-1/\theta-1}].$$

\textbf{Step 2: Form the Ratio $\phi_{A1}/\phi'_{A1}$.} Taking the ratio of the generator and its derivative simplifies to:
\begin{align*}
\frac{\phi_{A1}(t;\theta)}{\phi'_{A1}(t;\theta)} &= \frac{(t^{1/\theta} + t^{-1/\theta} - 2)^{\theta}}{(t^{1/\theta} + t^{-1/\theta} - 2)^{\theta-1}[t^{1/\theta-1} - t^{-1/\theta-1}]} \\
&= \frac{t^{1/\theta} + t^{-1/\theta} - 2}{t^{1/\theta-1} - t^{-1/\theta-1}}.
\end{align*}

Further algebraic simplification shows that this expression is equivalent to:
\[\frac{\phi_{A1}(t;\theta)}{\phi'_{A1}(t;\theta)} = \frac{t(t^{1/\theta} - 1)}{1+t^{1/\theta}}.\]

\textbf{Step 3: Change of Variables.} To evaluate the integral, we set $u = t^{1/\theta}$, which implies $t = u^{\theta}$ and $dt = \theta u^{\theta-1} du$. Substituting these into the integral from the corrected ratio in Step 2 gives:
\begin{align*}
I(\theta) = \int_0^1 \frac{\phi_{A1}(t;\theta)}{\phi'_{A1}(t;\theta)} dt &= \int_0^1 \frac{t(t^{1/\theta} - 1)}{1+t^{1/\theta}} dt \\
&= \int_0^1 \frac{u^{\theta}(u - 1)}{1+u} (\theta u^{\theta-1}) du \\
&= \theta \int_0^1 \frac{u^{2\theta-1}(u - 1)}{1 + u} du \\
&= \theta \int_0^1 \frac{u^{2\theta} - u^{2\theta-1}}{1 + u} du.
\end{align*}

\textbf{Step 4: Evaluate the Integral.} The integral can be solved using a standard identity for the digamma function, $\psi(\cdot)$, where:
$$\int_0^1 \frac{x^a - x^b}{1 + x} dx = \frac{1}{2}\left[\psi\left(\frac{a+2}{2}\right) - \psi\left(\frac{a+1}{2}\right) - \psi\left(\frac{b+2}{2}\right) + \psi\left(\frac{b+1}{2}\right)\right].$$

Setting $a = 2\theta$ and $b = 2\theta - 1$, the integral part becomes:
\begin{align*}
\int_0^1 \frac{u^{2\theta} - u^{2\theta-1}}{1+u} du &= \frac{1}{2}\left[\psi(\theta+1) - \psi\left(\theta+\frac{1}{2}\right) - \psi\left(\theta+\frac{1}{2}\right) + \psi(\theta)\right] \\
&= \frac{1}{2}\left[\psi(\theta+1) + \psi(\theta) - 2\psi\left(\theta+\frac{1}{2}\right)\right].
\end{align*}

Using the recurrence relation $\psi(\theta+1) = \psi(\theta) + 1/\theta$, this simplifies to:
$$\frac{1}{2}\left[(\psi(\theta) + \frac{1}{\theta}) + \psi(\theta) - 2\psi\left(\theta+\frac{1}{2}\right)\right] = \psi(\theta) - \psi\left(\theta+\frac{1}{2}\right) + \frac{1}{2\theta}.$$

Finally, we multiply by the leading factor of $\theta$ from Step 3:
$$I(\theta) = \theta\left[\psi(\theta) - \psi\left(\theta+\frac{1}{2}\right) + \frac{1}{2\theta}\right] = \theta\left[\psi(\theta) - \psi\left(\theta+\frac{1}{2}\right)\right] + \frac{1}{2}.$$

\textbf{Step 5: Final Expression for A1.} Substituting the correct integral value back into the formula for Kendall's $\tau$, we obtain the final expression:
\[\tau_{A1} = 1 + 4I(\theta) = 1 + 4\left(\theta\left[\psi(\theta) - \psi\left(\theta+\frac{1}{2}\right)\right] + \frac{1}{2}\right).\]
This implies:
\[\boxed{\tau_{A1} = 3 + 4\theta\left[\psi(\theta) - \psi\left(\theta+\frac{1}{2}\right)\right].}\]
\subsection{Derivation for the A2 Copula}
For the A2 copula, the generator is defined as
\[
\phi_{A2}(t;\theta) = \Bigl(\frac{1}{t}(1-t)^2\Bigr)^{\theta},\quad \theta\ge1.
\]
Differentiating via the log-derivative,
\[
\frac{\phi'_{A2}(t;\theta)}{\phi_{A2}(t;\theta)}
=\theta\left(\frac{d}{dt}\Bigl[2\ln(1-t)-\ln t\Bigr]\right)
=\theta\left(-\frac{2}{1-t}-\frac{1}{t}\right).
\]
Therefore,
\[
\frac{\phi_{A2}(t;\theta)}{\phi'_{A2}(t;\theta)}
=\frac{1}{\theta\left(-\frac{2}{1-t}-\frac{1}{t}\right)}
=\frac{t(t-1)}{\theta(t+1)}.
\]

Thus, Kendall's \(\tau\) is given by
\[
\tau_{A2} = 1 + 4\int_0^1 \frac{\phi_{A2}(t;\theta)}{\phi'_{A2}(t;\theta)}\,dt
= 1 + \frac{4}{\theta}\int_0^1 \frac{t(t-1)}{t+1}\,dt.
\]

\textbf{Step 1: Evaluate the Integral}
Define
\[
J = \int_0^1 \frac{t(t-1)}{t+1}\,dt.
\]
Since
\[
t(t-1) = t^2-t,
\]
we perform polynomial division of \(t^2-t\) by \(t+1\). Dividing, we obtain
\[
\frac{t^2-t}{t+1} = t-2+\frac{2}{t+1}.
\]
Thus,
\[
J = \int_0^1 \Bigl(t-2+\frac{2}{t+1}\Bigr)dt.
\]

\textbf{Step 2: Integrate Term-by-Term}
We compute each integral:
\[
\int_0^1 t\,dt = \left.\frac{t^2}{2}\right|_0^1 = \frac{1}{2},
\]
\[
\int_0^1 dt = 1,
\]
\[
\int_0^1 \frac{1}{t+1}\,dt = \left.\ln|t+1|\right|_0^1 = \ln2.
\]
Hence,
\[
J = \frac{1}{2} - 2\cdot 1 + 2\ln2 = \frac{1}{2} - 2 + 2\ln2 = -\frac{3}{2}+2\ln2.
\]

\textbf{Step 3: Final Expression for A2}
Substituting back into the expression for \(\tau_{A2}\), we have
\[
\boxed{
\tau_{A2}
= 1 - \frac{6-8\ln2}{\theta}\,.
}
\] 
\subsection{\texorpdfstring{Strict Monotonicity of $\tau_{A1}(\theta)$}{Strict Monotonicity of tau-A1(theta)}}
\label{sec:monotonicity}

\[\tau_{A1}(\theta) = 3 + 4\theta[\psi(\theta) - \psi(\theta + \frac{1}{2})], \quad \theta \geq 1.\]

\textbf{Claim.} $\tau_{A1}$ is strictly increasing on $[1,\infty)$; moreover
\[\tau_{A1}(1) = 8\ln 2 - 5 \approx 0.54518, \quad \lim_{\theta \to \infty} \tau_{A1}(\theta) = 1.\]

\textbf{Proof.}
Set
\[f(\theta) := \psi(\theta) - \psi\left(\theta + \frac{1}{2}\right).\]

A standard integral representation of the digamma function yields, for $x > 0$ and $a > 0$,
\[\psi(x) - \psi(x + a) = -\int_0^\infty \frac{1 - e^{-at}}{1 - e^{-t}} e^{-xt} dt.\]

With $a = \frac{1}{2}$ and $x = \theta$ we get
\[f(\theta) = -\int_0^\infty H(t) e^{-\theta t} dt, \quad H(t) := \frac{1 - e^{-t/2}}{1 - e^{-t}}, \quad t > 0.\]

Hence,
\begin{equation}
\tau_{A1}(\theta) = 3 - 4\theta \int_0^\infty H(t) e^{-\theta t} dt. \tag{A}
\end{equation}

\textbf{(i) Value at $\theta = 1$.}
Using $\psi(1) = -\gamma$ and $\psi\left(\frac{3}{2}\right) = \psi\left(\frac{1}{2}\right) + 2 = -\gamma - 2\ln 2 + 2$,
\[\tau_{A1}(1) = 3 + 4\left[\psi(1) - \psi\left(\frac{3}{2}\right)\right] = 3 + 4(2\ln 2 - 2) = 8\ln 2 - 5.\]

\textbf{(ii) Limit as $\theta \to \infty$.}
Near $t = 0$, $H(t)$ is continuous with $H(0) := \lim_{t \downarrow 0} \frac{1 - e^{-t/2}}{1 - e^{-t}} = \frac{1}{2}$.
Since $\theta e^{-\theta t}$ is an approximate identity on $[0,\infty)$,
\[\theta \int_0^\infty H(t) e^{-\theta t} dt \longrightarrow H(0) = \frac{1}{2}.\]

Taking this limit in (A) gives
\[\lim_{\theta \to \infty} \tau_{A1}(\theta) = 3 - 4 \cdot \frac{1}{2} = 1.\]

\textbf{(iii) Strict monotonicity on $[1,\infty)$.}
Differentiate (A):
\begin{equation}
\tau'_{A1}(\theta) = 4(f(\theta) + \theta f'(\theta)) = 4\int_0^\infty (\theta t - 1) H(t) e^{-\theta t} dt. \tag{B}
\end{equation}

We now show the right-hand side is strictly positive for every $\theta > 0$ (hence for $\theta \geq 1$).

First, observe that $H$ is strictly increasing on $(0,\infty)$. Indeed,
\begin{align*}
H'(t) &= \frac{\frac{1}{2}e^{-t/2}(1 - e^{-t}) - e^{-t}(1 - e^{-t/2})}{(1 - e^{-t})^2} \\
&= \frac{\frac{1}{2}e^{-t/2}(1 - e^{-t/2})^2}{(1 - e^{-t})^2} \\
&= \frac{1}{2} \frac{e^{-t/2}}{(1 + e^{-t/2})^2} > 0.
\end{align*}

Next, rewrite (B) by subtracting a zero term and integrating by parts in a monotone way. Since
\[\int_0^\infty (\theta t - 1)e^{-\theta t}dt = 0,\]
\[\int_0^\infty (\theta t - 1) H(t) e^{-\theta t}dt = \int_0^\infty (\theta t - 1) [H(t) - H(0)] e^{-\theta t}dt.\]

Write $H(t) - H(0) = \int_0^t H'(s) ds$, interchange integrals, and evaluate the inner integral:
\[\int_s^\infty (\theta t - 1)e^{-\theta t}dt = [-t e^{-\theta t}]_{t=s}^\infty = s e^{-\theta s}.\]

Therefore,
\[\int_0^\infty (\theta t - 1) H(t) e^{-\theta t}dt = \int_0^\infty s e^{-\theta s} H'(s) ds.\]

Since $s > 0$, $e^{-\theta s} > 0$, and $H'(s) > 0$ for all $s > 0$, the integrand is strictly positive on $(0,\infty)$, hence the integral is strictly positive. Combining with (B),
\[\tau'_{A1}(\theta) = 4\int_0^\infty s e^{-\theta s} H'(s) ds > 0 \quad (\theta > 0).\]

In particular, $\tau_{A1}$ is strictly increasing on $[1,\infty)$.

This completes the proof. $\square$

\textbf{Remark (explicit positive form).}
Using the closed form $H'(t) = \frac{1}{2} e^{-t/2}/(1 + e^{-t/2})^2$, the derivative can be written as
\[\tau'_{A1}(\theta) = 2\int_0^\infty \frac{s e^{-(\theta + \frac{1}{2})s}}{(1 + e^{-s/2})^2} ds > 0,\]
making strict positivity immediate.

\section{Identifiability Proofs for A1 and A2 Copulas}
\label{app:Appendix B}

\subsection{A1 Copula Identifiability}
\label{app:id_a1}

For the A1 family, Kendall's $\tau$ has the closed form
\[
  \tau_{A1}(\theta)
  = 3
    + 4\theta\!\left[
        \psi(\theta)
        - \psi\!\left(\theta+\tfrac{1}{2}\right)
      \right],
  \qquad \theta \ge 1,
\]
and we showed in Appendix~\ref{sec:monotonicity} that $\tau_{A1}(\theta)$ is
\emph{strictly increasing} on $[1,\infty)$. Hence if $\theta_1 \neq \theta_2$ then
$\tau_{A1}(\theta_1) \neq \tau_{A1}(\theta_2)$.

Therefore, the map $\theta \mapsto \tau_{A1}(\theta)$ is injective, and distinct parameter values induce distinct A1 copulas.
In particular,
$C(\cdot,\cdot;\theta_1)$ and $C(\cdot,\cdot;\theta_2)$ are distinct whenever
$\theta_1\neq\theta_2$, and the parameter $\theta$ is identifiable in the A1 family.

\subsection{A2 Copula Identifiability}
\label{app:id_a2}

For the A2 generator:
\[
  \phi_{A2}(t;\theta)
  = \left(\frac{(1-t)^2}{t}\right)^{\!\theta},
  \qquad \theta \ge 1,
\]
assume $\phi_{A2}(t;\theta_1) = \phi_{A2}(t;\theta_2)$ for all $t \in (0,1)$.
Taking logarithms yields
\[
  \theta_1\,\ln\!\left(\frac{(1-t)^2}{t}\right)
  = \theta_2\,\ln\!\left(\frac{(1-t)^2}{t}\right),
\]
and thus
\[
  (\theta_1-\theta_2)\,\ln\!\left(\frac{(1-t)^2}{t}\right)=0
  \qquad \text{for all } t\in(0,1).
\]
Choose, for example, $t=\tfrac{1}{2}$, for which
$\ln\!\left(\frac{(1-t)^2}{t}\right)=\ln\!\left(\tfrac{1}{2}\right)\neq 0$.
Then the above identity implies $\theta_1=\theta_2$. Hence the mapping
$\theta\mapsto \phi_{A2}(\cdot;\theta)$ is injective, and the parameter $\theta$
is identifiable in the A2 family.

\textbf{Conclusion.} Both proofs establish that
$\phi_{\theta_1} = \phi_{\theta_2} \implies \theta_1 = \theta_2$,
ensuring parameter identifiability for the A1 and A2 copula families.

\section{Consistency Proof for A1 and A2 Copulas}
\label{app:Appendix C}

\textbf{Regularity Conditions.}
For every copula family in $\{\text{Gumbel, Joe, A1, A2}\}$, we assume:

\textbf{1. Copula identifiability:}
The copula parameter is identifiable within each family, i.e.,
\[
  C(\cdot,\cdot;\theta_1)=C(\cdot,\cdot;\theta_2)
  \implies \theta_1=\theta_2.
\]
(See \citep{nelsen2006introduction} for the Gumbel and Joe copulas; for the
A1/A2 families we have given the proof in Appendix~\ref{app:Appendix B}.)

\textbf{2.} The generator $\phi_\theta$ is continuously differentiable in $\theta$.

\textbf{3. Feature Continuity and Injectivity:}
The vector of summary features
\[
  \mathbf{T}_n
  = \bigl(\tau_n,\;\rho_n,\;\lambda_{\text{upper},n},\;\lambda_{\text{lower},n},\;r_n\bigr)
\]
is well-defined for each $n$, and the population feature map
$\theta\mapsto \mathbf{T}(\theta)$ is continuous and injective on a compact
parameter set $\Theta$. Moreover, the empirical feature vector computed from
$n$ observations satisfies a standard consistency property (via classical
empirical-process arguments for rank-based copula functionals,
e.g.\ \citep{vandervaart1996weak}):
\[
  \mathbf{T}_n \xrightarrow{p} \mathbf{T}(\theta_0) \quad \text{as } n\to\infty,
\]
when the data are generated from the true parameter $\theta_0$ within the
specified copula family.

\begin{theorem}\label{thm:consistency}
Assume the regularity conditions above hold and further suppose that:

\textbf{1. Universal Approximation:}
There exists a neural network (NN) architecture that is dense in the space of
continuous functions over the compact feature domain; here, we assume that
$\Theta$ and the feature space $\mathcal{T}$ are compact, as required by
Hornik's theorem \citep{hornik1991approximation}.

\textbf{2. Training Density:}
As the number of training samples $N_{\text{train}}\to\infty$, the simulated
training design becomes dense over $\Theta$ within each copula family.

\textbf{3. Operational Regime:}
The number of real observations $n\to\infty$.

\textbf{4. Vanishing training/approximation error:}
The trained network $f_{\mathrm{NN}}(\cdot)$ approximates the population inverse
map $g^*(\cdot)$ (defined below) uniformly on the relevant compact feature
domain, with approximation/training error tending to zero as
$N_{\text{train}}\to\infty$.

\noindent Then the IGNIS estimator satisfies
\[
  \hat{\theta}_n \xrightarrow{p} \theta_0 \quad \text{as } n\to\infty.
\]
\end{theorem}

\begin{proof}
The proof proceeds in five steps.

\textbf{Step 1: Feature convergence.}
By the assumed consistency of the empirical summary features,
\[
  \mathbf{T}_n \xrightarrow{p} \mathbf{T}(\theta_0).
\]

\textbf{Step 2: Identifiability and the inverse map.}
Fix a copula family $C$. By injectivity of $\theta\mapsto \mathbf{T}(\theta)$
within $C$, there exists a well-defined inverse mapping on the image
$\mathbf{T}(\Theta)$:
\[
  g^*(\mathbf{T},C)=\theta, \qquad \mathbf{T}\in\mathbf{T}(\Theta),
\]
meaning that $g^*(\mathbf{T}(\theta),C)=\theta$ for all $\theta\in\Theta$.

Moreover, since $\Theta$ is compact and $\theta\mapsto \mathbf{T}(\theta)$ is
continuous and injective, the inverse map $g^*(\cdot,C)$ is continuous on
$\mathbf{T}(\Theta)$ (by the standard inverse mapping result for continuous
bijections on compact sets).

\textbf{Step 3: Universal approximation.}
By the universal approximation theorem \citep{hornik1991approximation}, for
any $\epsilon>0$ there exist network parameters $W$ such that
\[
  \sup_{(\mathbf{T},C)\,\in\,\mathbf{T}(\Theta)\times\mathcal{C}}
  \bigl|f_{\mathrm{NN}}(\mathbf{T},C;W)-g^*(\mathbf{T},C)\bigr|<\epsilon,
\]
where $\mathbf{T}(\Theta)$ is compact. (Here $C\in\mathcal{C}$ can be
represented via a finite-dimensional one-hot encoding, so the input domain is a
compact subset of Euclidean space.)

\textbf{Step 4: Training error control (large training set).}
Let $\widehat{W}$ denote the trained parameters obtained by minimizing an
empirical MSE over $N_{\text{train}}$ simulated training triples
$(\mathbf{T}_i,C_i,\theta_i)$:
\[
  \frac{1}{N_{\text{train}}}
  \sum_{i=1}^{N_{\text{train}}}
  \bigl(f_{\mathrm{NN}}(\mathbf{T}_i,C_i;\widehat{W})-\theta_i\bigr)^2.
\]
Under standard conditions for empirical risk minimization and well-specified
simulation-based training (see, e.g., \citep{white1989some}), the trained
network can be taken to approximate the population target map $g^*$ on the
relevant compact domain with error
$\delta_{N_{\text{train}}}\to 0$ as $N_{\text{train}}\to\infty$, i.e.,
\[
  \sup_{(\mathbf{T},C)\,\in\,\mathbf{T}(\Theta)\times\mathcal{C}}
  \bigl|f_{\mathrm{NN}}(\mathbf{T},C;\widehat{W})-g^*(\mathbf{T},C)\bigr|
  \le \delta_{N_{\text{train}}},
  \qquad \delta_{N_{\text{train}}}\to 0.
\]

\textbf{Step 5: Operational consistency.}
Define $\hat{\theta}_n := f_{\mathrm{NN}}(\mathbf{T}_n,C;\widehat{W})$ and note
that $\theta_0 = g^*(\mathbf{T}(\theta_0),C)$. Then
\begin{align*}
  |\hat{\theta}_n-\theta_0|
  &= \bigl|
       f_{\mathrm{NN}}(\mathbf{T}_n,C;\widehat{W})
       -g^*(\mathbf{T}(\theta_0),C)
     \bigr| \\
  &\le
    \underbrace{
      \bigl|
        f_{\mathrm{NN}}(\mathbf{T}_n,C;\widehat{W})
        -f_{\mathrm{NN}}(\mathbf{T}(\theta_0),C;\widehat{W})
      \bigr|
    }_{(a)}
    +
    \underbrace{
      \bigl|
        f_{\mathrm{NN}}(\mathbf{T}(\theta_0),C;\widehat{W})
        -g^*(\mathbf{T}(\theta_0),C)
      \bigr|
    }_{(b)}.
\end{align*}
Term~(b) is bounded by $\delta_{N_{\text{train}}}\to 0$ by Step~4.
For term~(a), since $f_{\mathrm{NN}}(\cdot,C;\widehat{W})$ is continuous on the
compact domain and $\mathbf{T}_n\xrightarrow{p}\mathbf{T}(\theta_0)$ by Step~1,
we obtain
\[
  \bigl|
    f_{\mathrm{NN}}(\mathbf{T}_n,C;\widehat{W})
    -f_{\mathrm{NN}}(\mathbf{T}(\theta_0),C;\widehat{W})
  \bigr| \xrightarrow{p} 0.
\]
Hence $|\hat{\theta}_n-\theta_0|\xrightarrow{p}0$, which proves
$\hat{\theta}_n\xrightarrow{p}\theta_0$.
\end{proof}

\subsection*{Practical Considerations}
In practice, the finite-sample performance of the IGNIS estimator can be
summarized via a standard decomposition of error sources:
\[
  \mathbb{E}\bigl[(\hat{\theta}_n-\theta_0)^2\bigr]
\;\lesssim\;
K_1\,n^{-1}
\;+\;
K_2\,N_{\text{train}}^{-1}
\;+\;
K_3\,\epsilon^2,
\]
where $K_1\,n^{-1}$ represents the estimation error due to finite $n$ (feature
noise), $K_2\,N_{\text{train}}^{-1}$ reflects finite training-set effects, and
$K_3\,\epsilon^2$ captures approximation error due to limited network
expressiveness. This decomposition is intended as a practical guideline:
overall performance is influenced by the sample size, the density of the
training data, and the expressiveness of the chosen neural network
architecture. Also, the feature vector is denoted by $\mathbf{T}_n$, and the
set of all possible feature vectors is the feature space $\mathcal{T}$.

\section{Pathological Properties of A1/A2 Copulas}
\label{app:Appendix D}

\noindent\textbf{Asymptotic regimes.}
We study two limits that isolate the main numerical barriers observed in likelihood-based estimation:
\begin{enumerate}
  \item \textbf{Density blowup (Barrier 1):} $t\to0^+$ with $\theta$ fixed, capturing boundary singularities of $\phi''(t;\theta)$.
  \item \textbf{Score/Hessian decay (Barriers 2--3):} $\theta\to\infty$ with fixed interior arguments $t\in(0,1)$, yielding
  score decay of orders $O(\theta^{-8})$ (A1) and $O(\theta^{-3})$ (A2), and corresponding Hessian decay
  $O(\theta^{-9})$ (A1) and $O(\theta^{-4})$ (A2).
\end{enumerate}

\subsection{Derivative Analysis}

\subsubsection{First and Second Derivatives of the A1 Generator}
For
\[
\phi_{A1}(t;\theta)=\bigl(t^{1/\theta}+t^{-1/\theta}-2\bigr)^{\theta},
\qquad
g(t)=t^{1/\theta}+t^{-1/\theta}-2,
\]
we have
\[
\phi'_{A1}(t)=\theta\,g(t)^{\theta-1}g'(t),
\qquad
\phi''_{A1}(t)=\theta(\theta-1)g(t)^{\theta-2}[g'(t)]^2+\theta g(t)^{\theta-1}g''(t),
\]
where
\[
g'(t)=\frac{1}{\theta}t^{1/\theta-1}-\frac{1}{\theta}t^{-1/\theta-1}
      =\frac{1}{\theta}t^{-1/\theta-1}\bigl(t^{2/\theta}-1\bigr),
\]
\[
g''(t)=\frac{1}{\theta}\left(\frac{1}{\theta}-1\right)t^{1/\theta-2}
      +\frac{1}{\theta}\left(\frac{1}{\theta}+1\right)t^{-1/\theta-2}.
\]

\subsubsection{First and Second Derivatives of the A2 Generator}
For
\[
\phi_{A2}(t;\theta)=\left(\frac{1-t}{t}\right)^{\theta}(1-t)^{\theta}
=(1-t)^{2\theta}t^{-\theta},
\]
direct differentiation gives
\begin{align*}
\phi'_{A2}(t)
&= -\theta\,(1-t)^{2\theta-1}\,t^{-\theta-1}\,(1+t),\\
\phi''_{A2}(t)
&= \theta\,(1-t)^{2\theta-2}\,t^{-\theta-2}
   \bigl[(\theta+1)+2(\theta-1)t+(\theta-1)t^2\bigr].
\end{align*}

\subsection{Barrier 1: Numerical Instability (Boundary Density Blowup)}

\begin{theorem}[Asymptotic singularity of $\phi''$ near $t\to0^+$]
\label{thm:barrier1}
As $t\to0^+$ with $\theta$ fixed,
\[
|\phi''_{A1}(t;\theta)| = \Theta(t^{-3}),
\qquad
|\phi''_{A2}(t;\theta)| = \Theta(t^{-\theta-2}).
\]
\end{theorem}

\begin{proof}
\textbf{A1.}
Recall
\[
\phi''_{A1}(t)=\theta(\theta-1)g^{\theta-2}[g']^2+\theta g^{\theta-1}g'',
\qquad
g(t)=t^{1/\theta}+t^{-1/\theta}-2.
\]
As $t\to0^+$, the dominant contribution to $g(t)$ is $t^{-1/\theta}$, hence
\[
g(t)\sim t^{-1/\theta}.
\]
Also $t^{2/\theta}\to0$, so
\[
g'(t)=\frac{1}{\theta}t^{-1/\theta-1}(t^{2/\theta}-1)\sim -\frac{1}{\theta}t^{-1/\theta-1},
\qquad
[g'(t)]^2\sim \frac{1}{\theta^2}t^{-2/\theta-2}.
\]
For $g''(t)$, the dominant term is the $t^{-1/\theta-2}$ component:
\[
g''(t)\sim \frac{1}{\theta}\Bigl(\frac{1}{\theta}+1\Bigr)t^{-1/\theta-2}.
\]
Substituting:
\begin{align*}
\theta(\theta-1)g^{\theta-2}[g']^2
&\sim \theta(\theta-1)\,(t^{-1/\theta})^{\theta-2}\,\frac{1}{\theta^2}t^{-2/\theta-2}
= \frac{\theta-1}{\theta}\,t^{-3},\\
\theta g^{\theta-1}g''
&\sim \theta\,(t^{-1/\theta})^{\theta-1}\,\frac{1}{\theta}\Bigl(\frac{1}{\theta}+1\Bigr)t^{-1/\theta-2}
= \Bigl(\frac{1}{\theta}+1\Bigr)t^{-3}.
\end{align*}
Therefore $\phi''_{A1}(t)\sim \bigl(\frac{\theta-1}{\theta}+\frac{1}{\theta}+1\bigr)t^{-3}=2t^{-3}$, proving $\Theta(t^{-3})$.

\textbf{A2.}
Using the closed form,
\[
\phi''_{A2}(t)=\theta(1-t)^{2\theta-2}t^{-\theta-2}\bigl[(\theta+1)+2(\theta-1)t+(\theta-1)t^2\bigr].
\]
As $t\to0^+$, $(1-t)^{2\theta-2}\to1$ and the bracketed polynomial tends to $\theta+1$, so
\[
\phi''_{A2}(t)\sim \theta(\theta+1)t^{-\theta-2},
\]
which is $\Theta(t^{-\theta-2})$.
\end{proof}

\subsection{Barrier 2: Vanishing Gradients (Score Decay and Likelihood Plateaus)}

\begin{theorem}[Score-decay rates]
\label{thm:barrier2_decay}
Let $\ell(\theta)=\sum_{i=1}^n \log c(u_i,v_i;\theta)$ be the log-likelihood.
For fixed interior arguments (i.e., away from $t\to0$ and $t\to1$) as $\theta\to\infty$,
\[
|\partial_\theta \ell(\theta)|
=
\begin{cases}
O\bigl(n\,\theta^{-8}\bigr), & \text{A1},\\
O\bigl(n\,\theta^{-3}\bigr), & \text{A2}.
\end{cases}
\]
\end{theorem}

\begin{proof}
The score $\partial_\theta \ell(\theta)$ is a sum of per-observation contributions
$\partial_\theta \log c(u_i,v_i;\theta)$. For Archimedean copulas under the convention $\phi:(0,1]\to[0,\infty)$ with
$C(u,v;\theta)=\phi^{-1}(\phi(u;\theta)+\phi(v;\theta);\theta)$, the log-density
$\log c(u_i,v_i;\theta)$ can be written in terms of $\phi',\phi'',\phi'''$ evaluated
at $u_i,v_i$ and at the copula value $w_i=C(u_i,v_i;\theta)$ (via the standard identity
$c(u,v;\theta)=-\phi''(w;\theta)\phi'(u;\theta)\phi'(v;\theta)/\{\phi'(w;\theta)\}^3$).
The score therefore involves both (i) explicit $\theta$-dependence of $\phi',\phi''$
and (ii) implicit $\theta$-dependence through $w_i=w_i(\theta)$.

In the limit $\theta\to\infty$ with fixed interior $t\in(0,1)$, one can expand the relevant
building blocks in powers of $1/\theta$. For A1, using $L=\ln t$,
\[
t^{\pm 1/\theta}=e^{\pm L/\theta}
=1\pm \frac{L}{\theta}+\frac{L^2}{2\theta^2}\pm\frac{L^3}{6\theta^3}+\frac{L^4}{24\theta^4}+O(\theta^{-5}),
\]
which implies
\[
g(t)=t^{1/\theta}+t^{-1/\theta}-2=\frac{L^2}{\theta^2}+\frac{L^4}{12\theta^4}+O(\theta^{-6}).
\]
Substituting these expansions into the exact score expression and collecting powers of $1/\theta$
reveals cancellations of the leading orders; the first non-canceling contribution
appears at order $\theta^{-8}$ (A1) and $\theta^{-3}$ (A2) for interior arguments.

Thus per observation $\partial_\theta \log c=O(\theta^{-8})$, and summing over $n$ gives
$|\partial_\theta\ell(\theta)|=O(n\theta^{-8})$.

For A2, an analogous expansion using the explicit closed forms for $\phi'_{A2}$ and $\phi''_{A2}$
and the implicit $\theta$-dependence of the copula value $w_i$  yields per-observation decay
$O(\theta^{-3})$, and hence $|\partial_\theta\ell(\theta)|=O(n\theta^{-3})$.
\end{proof}

\begin{theorem}[Definition and computation of $\theta_{\mathrm{crit}}$ used in Figures]
\label{thm:barrier2_theta_crit}
Assume the leading magnitude of the score satisfies the practical approximation
\[
|\partial_\theta \ell(\theta)| \approx C\,n\,\theta^{-k},
\]
with $(k,C)=(8,C_1)$ for A1 and $(k,C)=(3,C_2)$ for A2. For a gradient tolerance
$\varepsilon_{\mathrm{grad}}$, define $\theta_{\mathrm{crit}}$ by the flatness condition
\[
|\partial_\theta \ell(\theta)|\le \varepsilon_{\mathrm{grad}}.
\]
Then
\[
\theta_{\mathrm{crit}}=\Bigl(\frac{C\,n}{\varepsilon_{\mathrm{grad}}}\Bigr)^{1/k}.
\]
With $n=1000$, $\varepsilon_{\mathrm{grad}}=10^{-6}$, $C_1=0.02$ and $C_2=0.002$,
\[
\theta_{\mathrm{crit}}^{\mathrm{A1}}=\Bigl(\frac{0.02\cdot 1000}{10^{-6}}\Bigr)^{1/8}
=(2\times 10^7)^{1/8}\approx 8.17,
\]
\[
\theta_{\mathrm{crit}}^{\mathrm{A2}}=\Bigl(\frac{0.002\cdot 1000}{10^{-6}}\Bigr)^{1/3}
=(2\times 10^6)^{1/3}\approx 125.99\approx 126.
\]
\end{theorem}

\begin{proof}
Solve $C n \theta^{-k}=\varepsilon_{\mathrm{grad}}$ for $\theta$ to obtain
$\theta=(Cn/\varepsilon_{\mathrm{grad}})^{1/k}$. The numerical values follow by direct substitution.
The prefactors $C_1,C_2$ are calibration constants used for visualization and depend on scaling and
the particular dataset/normalization used in the plotted score proxy.
\end{proof}

\paragraph{Y-axis definition in Figures \ref{fig:A1_likelihood_surface}--\ref{fig:A2_likelihood_surface}.}
The plotted quantity is a \emph{normalized score magnitude proxy}:
we display a nonnegative score magnitude (e.g., $|\partial_\theta\ell(\theta)|$ or a closely related
quantity used in the computation) scaled to lie in $[0,1]$ over the plotted $\theta$-range
by dividing by its maximum value on that plotting grid.

\subsection{Barrier 3: Hessian Decay}

\begin{theorem}[Hessian decay rates and practical “numerical zero” thresholds]
\label{thm:barrier3}
As $\theta\to\infty$ with interior arguments fixed,
\[
|\partial_\theta^2 \ell(\theta)|
=
\begin{cases}
O\bigl(n\,\theta^{-9}\bigr), & \text{A1},\\
O\bigl(n\,\theta^{-4}\bigr), & \text{A2}.
\end{cases}
\]
A practical criterion for Newton-type instability is when the scalar Hessian becomes
numerically negligible relative to unit-scale quantities, e.g.,
\(
|\partial_\theta^2\ell(\theta)| \lesssim \varepsilon_{\mathrm{mach}}
\)
in double precision with $\varepsilon_{\mathrm{mach}}\approx 2.22\times 10^{-16}$.
Under this \emph{practical} criterion, for $n=1000$ one obtains thresholds
\[
\theta \gtrsim (n/\varepsilon_{\mathrm{mach}})^{1/9}\approx 1.2\times 10^2 \quad \text{(A1)},
\qquad
\theta \gtrsim (n/\varepsilon_{\mathrm{mach}})^{1/4}\approx 4.6\times 10^4 \quad \text{(A2)}.
\]
\end{theorem}

\begin{proof}
If per observation $\partial_\theta \log c = O(\theta^{-k})$, then differentiating once more gives
$\partial_\theta^2 \log c = O(\theta^{-(k+1)})$.
Summing $n$ observations yields $|\partial_\theta^2\ell(\theta)|=O(n\theta^{-(k+1)})$.
For A1, $k=8$ so $k+1=9$; for A2, $k=3$ so $k+1=4$.

For the practical thresholds, impose $n\theta^{-(k+1)}\approx \varepsilon_{\mathrm{mach}}$ and solve
$\theta \approx (n/\varepsilon_{\mathrm{mach}})^{1/(k+1)}$.
We emphasize that this is a \emph{numerical-negligibility} criterion (relevant for optimization),
not an IEEE underflow-to-zero statement.
\end{proof}

\begin{figure}[ht]
\centering
\begin{subfigure}{0.48\textwidth}
    \centering
    \includegraphics[width=\textwidth]{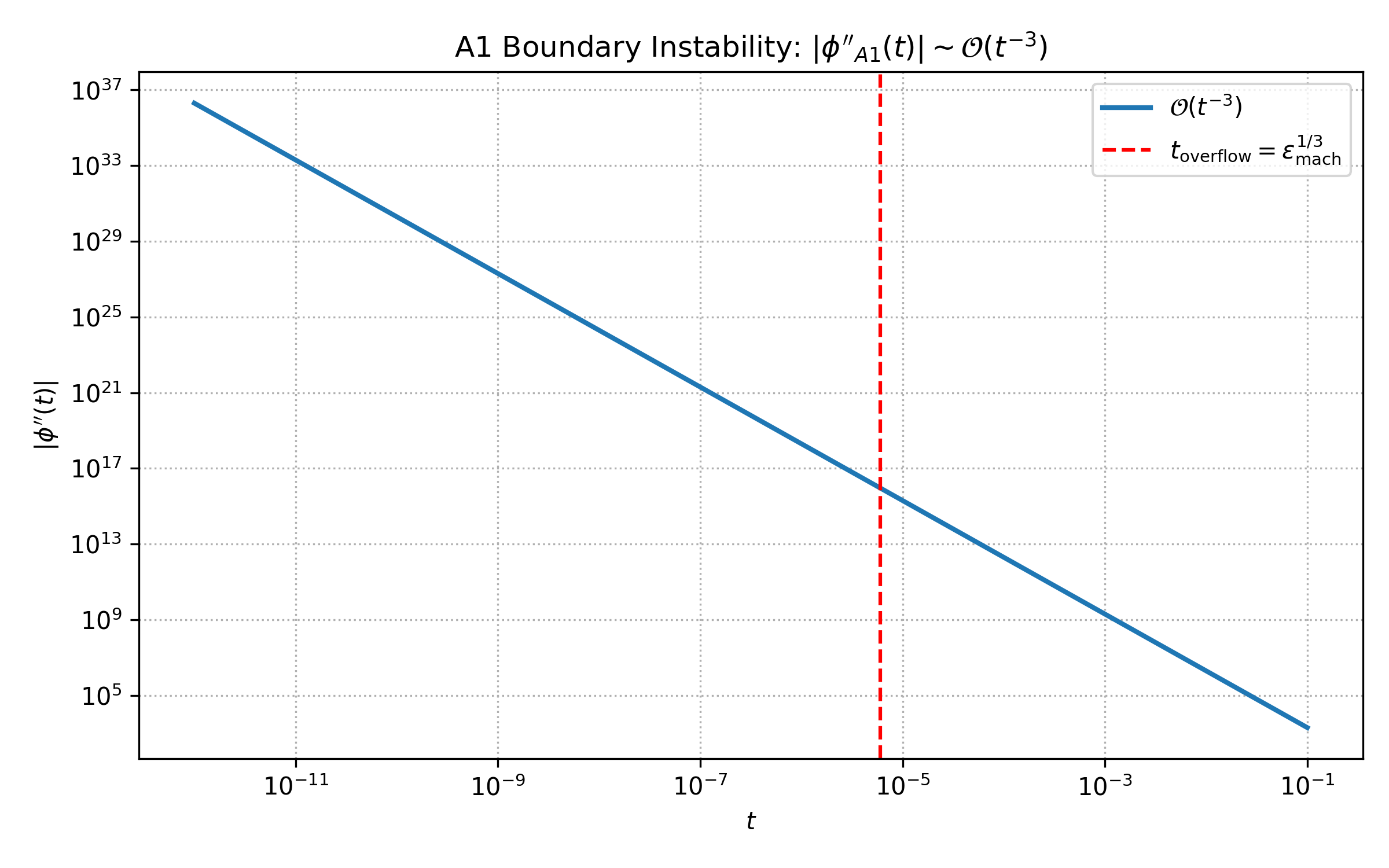}
    \caption{A1 copula boundary instability}
    \label{fig:A1_boundary_instability}
\end{subfigure}
\hfill
\begin{subfigure}{0.48\textwidth}
    \centering
    \includegraphics[width=\textwidth]{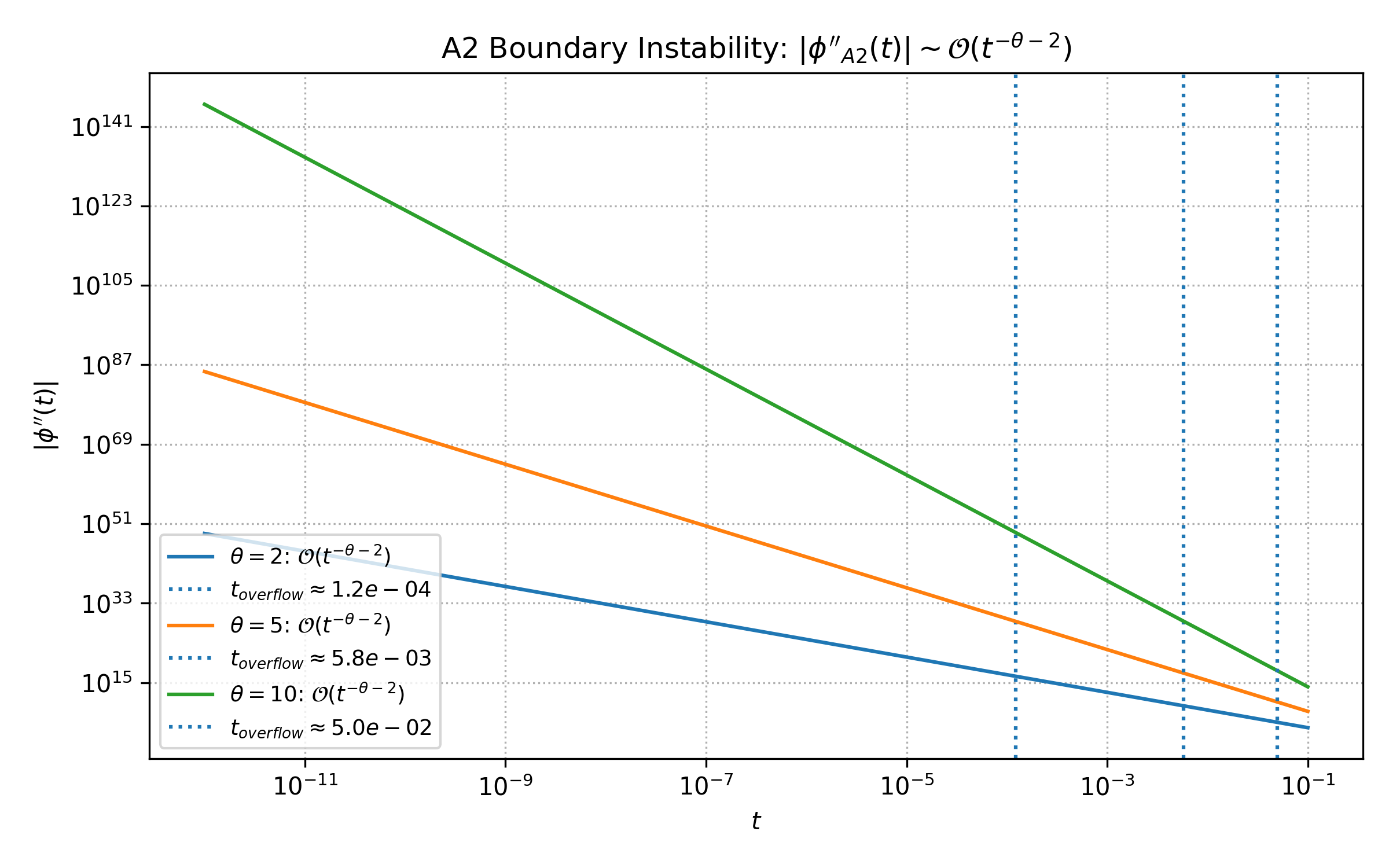}
    \caption{A2 copula boundary instability}
    \label{fig:A2_boundary_instability}
\end{subfigure}

\vspace{0.5cm}

\begin{subfigure}{0.48\textwidth}
    \centering
    \includegraphics[width=\textwidth]{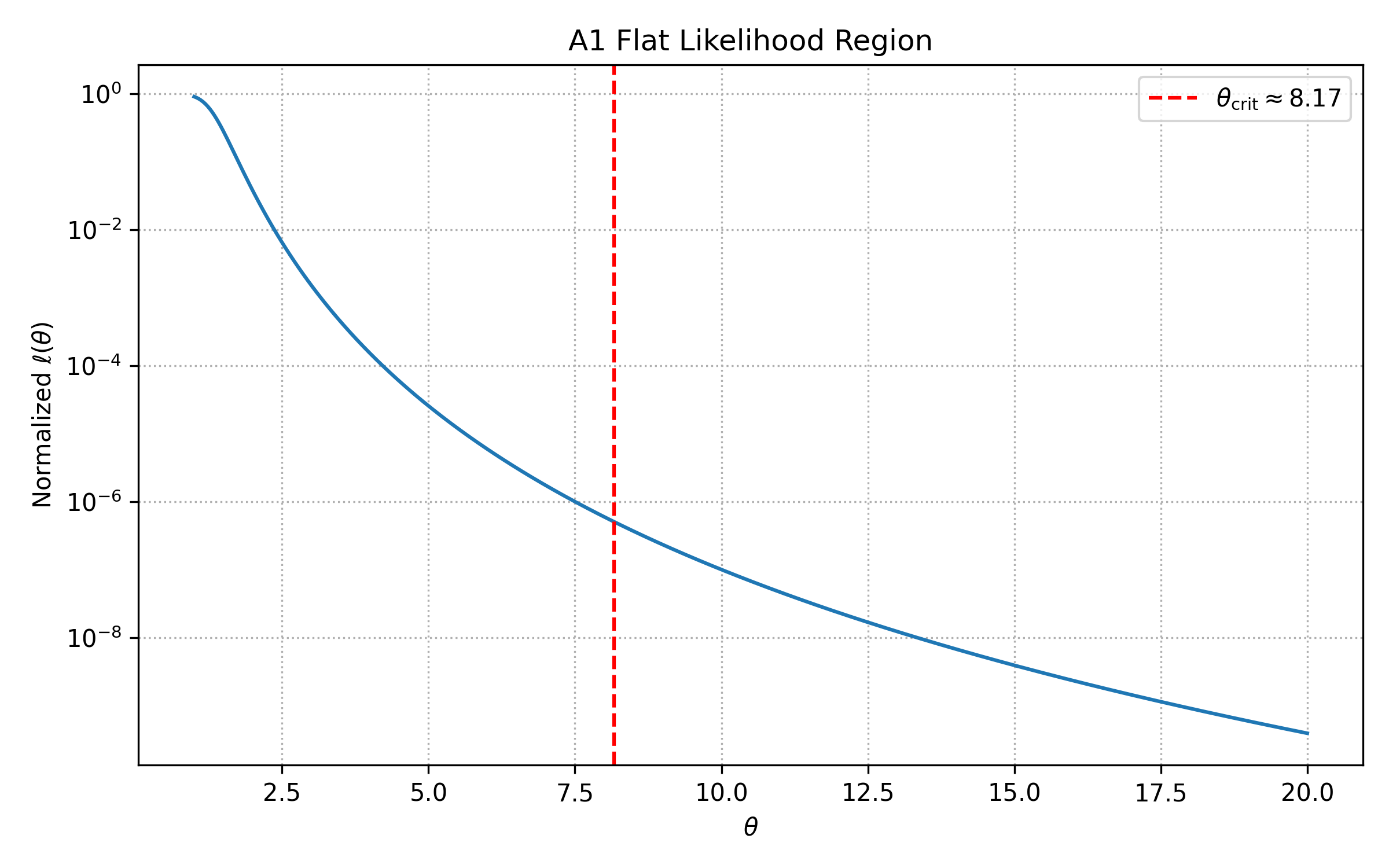}
    \caption{A1 flat-likelihood region (normalized score proxy vs.\ $\theta$)}
    \label{fig:A1_likelihood_surface}
\end{subfigure}
\hfill
\begin{subfigure}{0.48\textwidth}
    \centering
    \includegraphics[width=\textwidth]{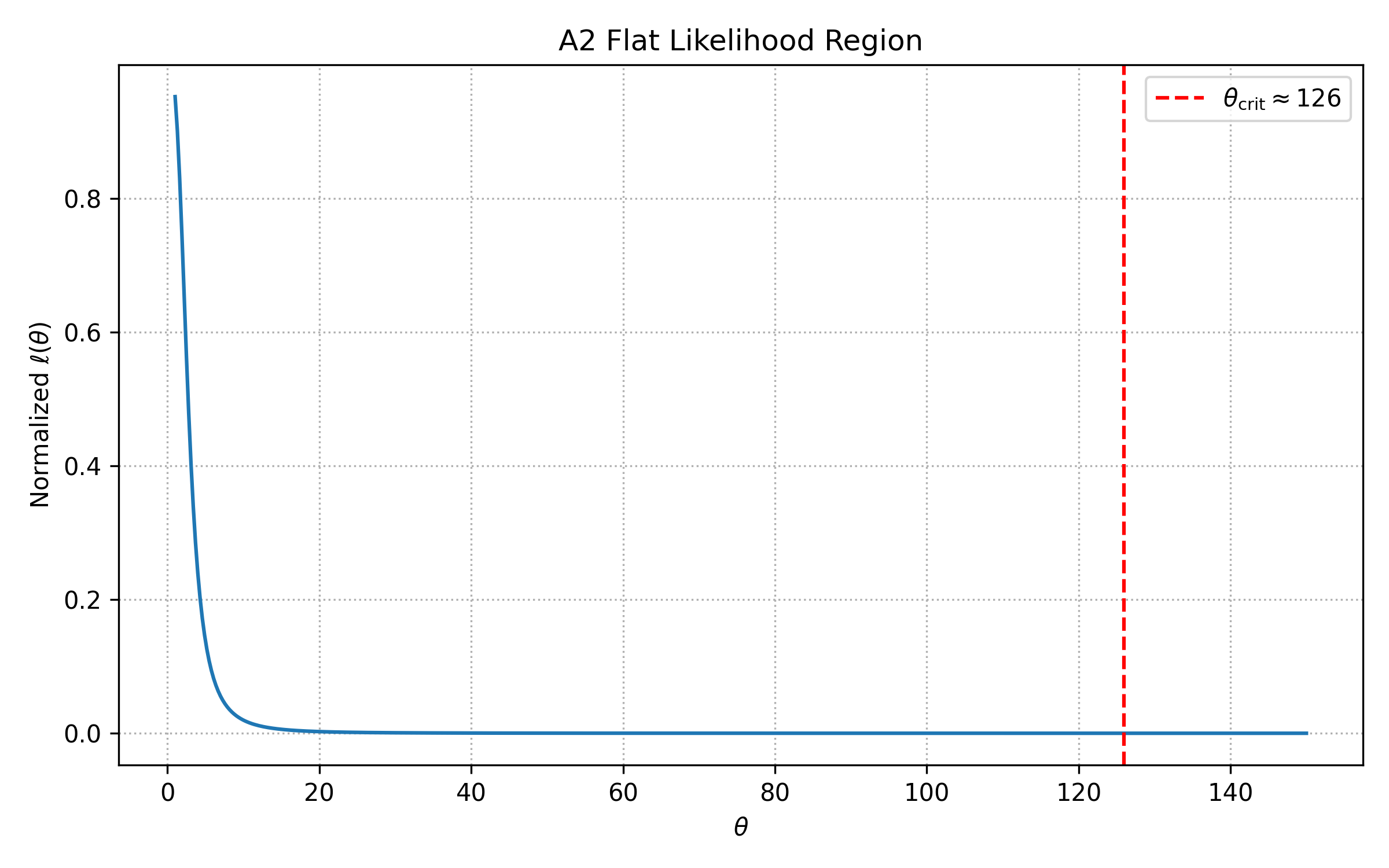}
    \caption{A2 flat-likelihood region (normalized score proxy vs.\ $\theta$)}
    \label{fig:A2_likelihood_surface}
\end{subfigure}

\vspace{0.5cm}

\begin{subfigure}{0.48\textwidth}
    \centering
    \includegraphics[width=\textwidth]{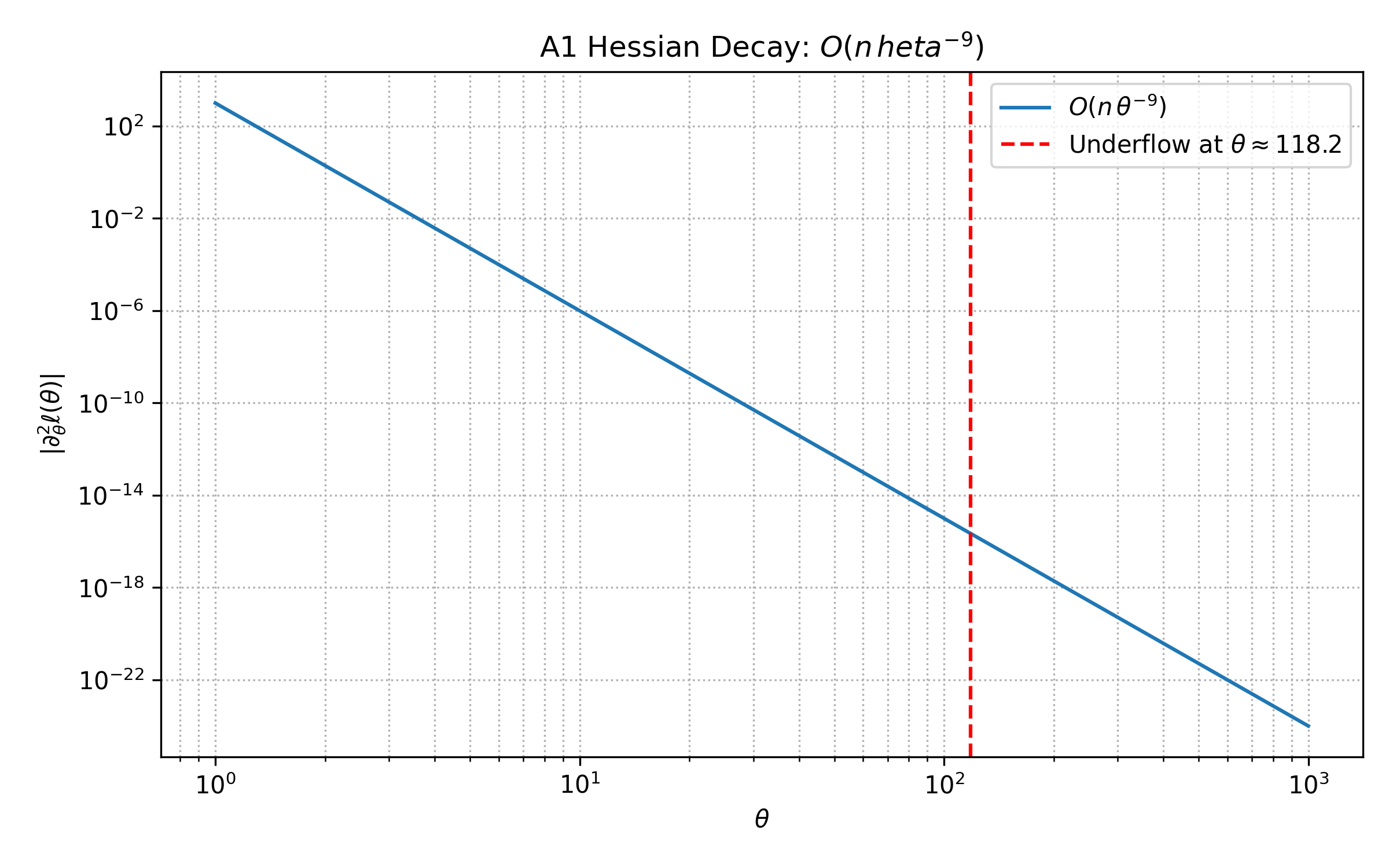}
    \caption{A1 Hessian decay/conditioning}
    \label{fig:A1_hessian_condition}
\end{subfigure}
\hfill
\begin{subfigure}{0.48\textwidth}
    \centering
    \includegraphics[width=\textwidth]{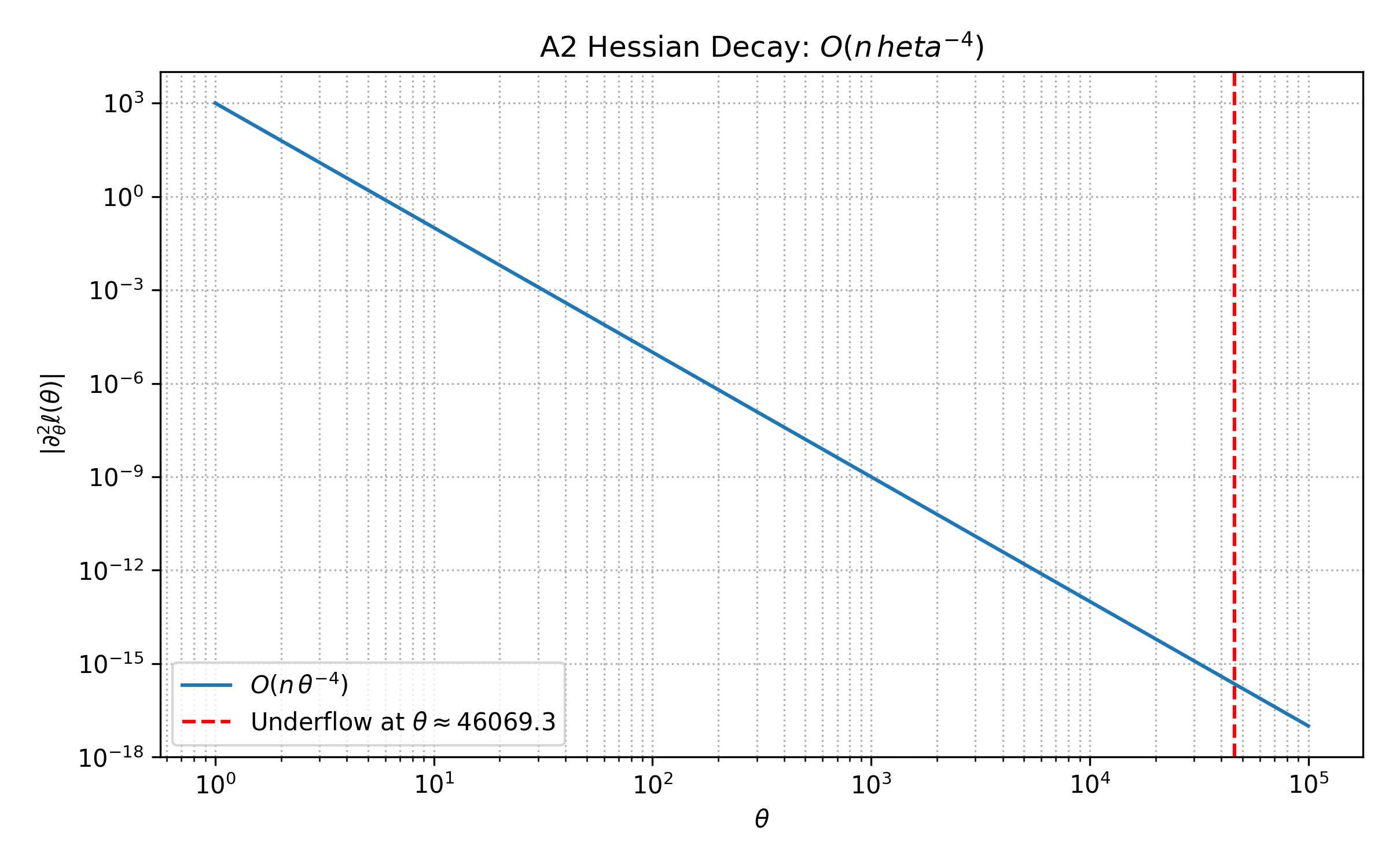}
    \caption{A2 Hessian decay/conditioning}
    \label{fig:A2_hessian_condition}
\end{subfigure}
\caption{Numerical challenges in copula estimation: boundary instabilities, flat likelihood regions, and Hessian decay/conditioning for A1 and A2 copulas. Dashed lines mark machine-precision based numerical-negligibility thresholds relevant for optimization stability.}
\label{fig:copula_estimation_challenges}
\end{figure}

\paragraph{Visualization of numerical barriers.}
Figure~\ref{fig:copula_estimation_challenges} provides a visual summary of the three numerical barriers
described above: boundary instability as $t\to0^+$ (Barrier~1), flat score/likelihood regions induced by
vanishing gradients as $\theta$ grows (Barrier~2), and Hessian decay/poor conditioning for Newton-type updates
(Barrier~3).

\end{document}